\documentclass[10pt,journal,compsoc]{IEEEtran}

\usepackage{amsmath,amssymb,amsfonts,bm, amsthm}
\usepackage{pgfplots}
\usepackage{filecontents}
\usepackage{multirow}
\usepackage[draft]{todonotes}   
\usepackage{multirow}
\usepackage{tabularx}
\usepackage{setspace}
\usepackage{listings}
\usepackage{comment}
\usepackage{lipsum,multicol}
\usepackage{color,soul}
\usepackage{subfig}
\usepackage{wrapfig}
\usepackage{algorithm} 
\usepackage[noend]{algpseudocode}
\usepackage{varwidth}
\usepackage{balance}

\usepackage{mathtools}
\usepackage{makecell}

\theoremstyle{definition}

\usepackage{url}

\newcommand{\new}[1]{{\leavevmode\color{black}#1}}
\newcommand{\old}[1]{}

\begin{document}

\title{FLSys: Toward an Open Ecosystem for Federated Learning Mobile Apps}

\author{
\IEEEauthorblockN{
	Xiaopeng Jiang \IEEEauthorrefmark{1}
	Han Hu \IEEEauthorrefmark{1}
	Thinh On\IEEEauthorrefmark{1}
	Phung Lai\IEEEauthorrefmark{1}
	Vijaya Datta Mayyuri \IEEEauthorrefmark{2}
	An Chen \IEEEauthorrefmark{2} }
	\\
\IEEEauthorblockN{
	Devu M. Shila\IEEEauthorrefmark{3}
	Adriaan Larmuseau\IEEEauthorrefmark{3}
	Ruoming Jin\IEEEauthorrefmark{4}
	Cristian Borcea\IEEEauthorrefmark{1}
	NhatHai Phan\IEEEauthorrefmark{1}
}
\\
\IEEEauthorblockN{
	New Jersey Institute of Technology\IEEEauthorrefmark{1}
	Qualcomm Incorporated\IEEEauthorrefmark{2}\\
}
\IEEEauthorblockN{
	Unknot.id\IEEEauthorrefmark{3}
	Kent State University\IEEEauthorrefmark{4}
}

\IEEEauthorblockA{		
		Email:\{xj8,hh255,to58,tl353,borcea,phan\}@njit.edu, \{vmayyuri,anc\}@qualcomm.com\\
		}
\IEEEauthorblockA{
		\{devums,adriaan\}@unknot.id,
		rjin1@kent.edu 
		}
}
\markboth{IEEE Transactions on Mobile Computing}%
{Shell \MakeLowercase{\textit{et al.}}: Bare Demo of IEEEtran.cls for Computer Society Journals}

\IEEEtitleabstractindextext{
\begin{abstract}

This article presents the design, implementation, and evaluation of \textbf{FLSys}, a mobile-cloud federated learning (FL) system, which can be a key component for an open ecosystem of FL models and apps. FLSys is designed to work on smart phones with mobile sensing data. It balances model performance with resource consumption, tolerates communication failures, and achieves scalability. In FLSys, different DL models with different FL aggregation methods can be trained and accessed concurrently by different apps. Furthermore, FLSys provides advanced privacy preserving mechanisms and a common API for third-party app developers to access FL models. FLSys adopts a modular design and is implemented in Android and AWS cloud. We co-designed FLSys with a human activity recognition (HAR) model. HAR sensing data was collected in the wild from 100+ college students during a 4-month period. We implemented HAR-Wild, a CNN model tailored to mobile devices, with a data augmentation mechanism to mitigate the problem of non-Independent and Identically Distributed data. A sentiment analysis model is also used to demonstrate that FLSys effectively supports concurrent models. This article reports our experience and lessons learned from conducting extensive experiments using simulations, Android/Linux emulations, and Android phones that demonstrate FLSys achieves good model utility and practical system performance.

\end{abstract}

}

\maketitle

\IEEEraisesectionheading{\section{Introduction} 
\label{sec:intro}
}
\IEEEPARstart{F}{ederated} Learning (FL)~\cite{bonawitz2019towards} has the potential to bring deep learning (DL) on mobile devices, while preserving user privacy during model training. FL balances model performance and user privacy through three design features. First, each device trains a local model on its raw data. Second, the gradients of the local models from multiple users are sent to a server for aggregation to compute a global model that is more accurate than individual local models. Third, the server shares the global model with all users. During this federated training, the raw data from individual users never leave their devices. A wide range of mobile apps, e.g., predicting or classifying health conditions based on mobile sensing data, can benefit from running DL models on smart phones using FL, which offers privacy-preserving global training that incentivizes user participation. 

Despite progress on theoretical aspects and algorithm/model design for FL~\cite{DBLP:journals/corr/abs-1812-06127,sarkar2020fedfocal,zhao2018federated,9141436,10.1117/12.2519621}, the lack of a publicly available FL system targeting mobile devices has precluded the widespread adoption of FL models on smart phones, even though such models can enable novel mobile apps that apply DL on mobile data (many times collected from sensors on the phones) in a privacy-preserving manner. Furthermore, this
has also limited our understanding of how real-world applications can benefit from FL. 
Most of existing FL systems are either unavailable for the research and practice communities (e.g., Google~\cite{bonawitz2019towards}, FedVision \cite{liu2020fedvision}), under development~\cite{he2020fedml}, or do not support mobile devices~\cite{FATE}. Well-developed open systems enabling on-device training~\cite{pysyft,beutel2021flower} do not provide support for third-party app development and do not consider the constraints of mobile devices.
Most of the existing FL studies are based on simulations~\cite{mugunthan2020privacyfl,DBLP:journals/corr/abs-1812-06127,sarkar2020fedfocal,zhao2018federated,9141436,10.1117/12.2519621}, which may lead to an oversimplified view of the applicability of FL models in real-world. In the meantime, although demonstrated in several scenarios such as keyboard typing prediction~\cite{yang2018applied}, FL lacks real-world applications, which can drive the design of FL systems. Indeed, real-world benchmarks for FL are pivotal to help shape the developments of FL systems~\cite{li2020federated}. 

In this article, we take a unique application-system co-design approach to design, build, and evaluate an FL system. Our system design is informed by
a critical mobile app, which illustrates a large category of apps that use DL on mobile sensing data: human activity recognition (HAR) on smart phones, which is important for industry, public health, and research. Simply speaking, mobile apps using HAR can harness recognized human physical activities using data collected from phone sensors. HAR is a representative FL app on smart phones that needs privacy-sensitive mobile sensing data collected in the wild in order to work effectively. Most FL papers use simulations with data collected offline and/or in controlled environment~\cite{ignatov2018real,murad2017deep,hernandez2019human,kwapisz2011WISDM,anguita2013ucihar,chavarriaga2013opportunity,chen2016lstm}, which do not work well in real-world. Furthermore, they do not consider the inter-play between concurrent data collection, training, and inference on model utility and resource consumption on the phones. From an industry point of view, accurate HAR can help the smart phone manufacturers to intelligently allocate resources and extend battery life. Users' behaviors, revealed by HAR data collected over long periods of time, may be privacy-sensitive, especially when location data is collected in addition to inertial measurement unit (IMU) data. Furthermore, collecting user data at a central server for training may violate recent privacy regulations (e.g., GDPR).
In general, the privacy-sensitive nature of mobile sensing data, which may also include photos and videos, makes HAR ideal for studying the design of FL systems.
 
In addition to HAR, we analyzed other real-life applications \cite{DBLP:journals/corr/abs-1912-04977,bonawitz2019towards,yang2019federated,yang2018applied,liu2020fedvision} to inform the system design. A list of important questions emerges, and many of them are not addressed in existing FL system designs~\cite{bonawitz2019towards,yang2018applied,he2020fedml,yang2019federated} that largely ignore the constraints of mobile devices:
How can we balance FL model performance with resource constraints on the phones? How can we ensure the training conducted on phones is completed on time, despite limited resources, i.e., computation power and battery life? 
How can the server achieve seamless scalability and accurate model aggregation in the presence of large and variable numbers of users who typically train different models and how can the system simultaneously cope with potential communication failures (e.g., connectivity lost on the phone)?
After a global model is shared with the phones, how can a third-party DL app utilize this model? How does the system support different types of advanced privacy preserving mechanisms?

\textbf{Key Contributions.} 
\new{This is the first article to provide a comprehensive description of the design, implementation, and evaluation of an FL system for smart phones, \textbf{FLSys}. The two main challenges for an FL system on phones are concurrent management of multiple FL activities under resource constraints and frequent disconnections due to networking and battery issues. These two challenges are not considered by any existing FL system. To solve them, we propose a novel system architecture that provides (1) a unified system to manage resources on the phone in the presence of multiple models, third-party apps using these models, and data collectors for these models; and (2) an asynchronous protocol to manage the FL process in the presence of disconnections. The FLSys components on smart phones manage training, inference, data collection/preprocessing, and privacy to balance model utility with resource consumption, while tolerating disconnections.

Furthermore, the engineering of an effective and efficient FLSys prototype on Android and AWS and its evaluation with data collected in the wild is also a major novel contribution of this article. No such system is currently available to the research community. While implemented in Android and AWS, FLSys has a general system design and API that can be extended to other mobile OSs and cloud platforms. 

At a more specific level, there are four novel contributions in the system architecture that combine solutions in machine learning, fault-tolerance, software engineering, and cloud systems. First, FLSys balances model performance, privacy and resource consumption on-demand through data collection and training configurations, such as sampling rate, model structure, hyper-parameters, and differential privacy (DP) mechanisms. Second, FLSys uses an asynchronous protocol between the server and the phones to handle phone failures to participate in training due to resource constraints or disconnections, while maintaining good model performance. This protocol allows the devices to self-select for training when they have enough data and resources and allows the sever to operate correctly in the presence of communication failures with the phones. Third, FLSys enables an ecosystem of third-party apps and models, as well as the ability to use different aggregators, data collectors/preprocessors, and DP-based privacy mechanisms through its modular design. FLSys provides a common API for third-party apps to retrieve inference results from different DL models, while efficiently managing resource consumption and contention. FLSys also flexibly supports different types of DP mechanisms, both on the mobile devices and in the cloud to protect user privacy against an honest-but-curious server. 
Fourth, in FLSys, different aggregation algorithms and training policies can be deployed selectively as modules in the cloud using function as a service (FaaS) support, which makes operating FL more cost-efficient. We also leverage FaaS and cloud storage solutions to engineer a scalable FL server.

Another novel contribution of this article is the HAR model that we designed and built to test FLSys, which is tailored to work efficiently on resource-constrained phones with non independent and identically distributed (non-IID) data. For HAR experiments on FLSys, we collected data from 100+ college students in two areas during a 4-month period. The students used their own Android phones, and their daily-life activities were not constrained in any way by our experiment. Data collected on mobile devices are non-IID, which affects FL-trained models~\cite{DBLP:journals/corr/abs-1912-04977}. 
We have evaluated a variety of HAR models with both centralized and federated training, and designed \textbf{HAR-Wild}, a Convolution Neural Network (CNN) model with a data augmentation mechanism to mitigate the non-IID problem. HAR-Wild was also designed to have a small memory footprint, which is appropriate for resource-constrained devices. To showcase the ability of FLSys to work with different FL models, we also built and evaluated a natural language sentiment analysis (SA) model on a dataset with 46,000+ tweets from 436 users.\footnote{The dataset was downloaded and evaluated by the NJIT team.}} 

We carried out a comprehensive evaluation of FLSys together with HAR-Wild and SA to quantify the model utility and the system feasibility in real life conditions. This article is the first in the literature to share an extensive FL evaluation on smart phones, using an end-to-end mobile-cloud FL system and mobile data collected in the wild.
We performed the evaluation under three training settings: 1) centralized training, 2) simulated FL with advanced privacy preserving mechanisms, and 3) Android FL. Centralized training provides an upper bound on model accuracy and is used to compare our HAR-Wild model with baseline approaches. 
The results demonstrate that HAR-Wild outperforms the baseline models in terms of accuracy. Furthermore, the federated HAR-Wild performance using simulations (TensorFlow and DL4J~\footnote{https://deeplearning4j.org/}), Android emulations, and Android phone experiments is close to the upper bound performance achieved by the centralized model. The results on smart phones demonstrate that FLSys can perform communication and training tasks within the allocated time and resource limits, while the FL server is able to handle a variable number of users. 
Finally, micro-benchmarks on Android phones show FLSys with HAR-Wild and SA are practical in terms of training and inference time, as well as memory and battery consumption.  

The rest of the article is organized as follows. Section~\ref{sec:related} discusses related work.
Section~\ref{sec:design} explains the design of FLSys, while Section~\ref{sec:implementation} describes its prototype implementation. Section~\ref{sec:model} presents the HAR model and data. Section~\ref{sec:results} shows the experimental results. The article concludes in Section~\ref{sec:conclusion} with lessons learned and future work.
	

\section{Related Work}
\label{sec:related}

This section reviews related work for FL systems, heterogeneity issues in FL federated training, and HAR models.  

\subsection{Federated Learning Systems}
FL can be categorized into Horizontal FL, Vertical FL, and Federated Transfer Learning (FTL)~\cite{yang2019federated}. In Horizontal FL, data are partitioned by device user Ids, such that users share the same feature space~\cite{yang2019federated}. In Vertical FL, different organizations have a large overlapping user space with different feature spaces. These organizations aim at jointly training a model to predict the same model outcomes, without sharing their data. In FTL, the datasets of these organizations differ in both the user space and the feature space. 
In Vertical FL and FTL, different organizations need to align their common users and exchange intermediate results by applying encryption techniques~\cite{9006000}. The server cannot just average the gradients, but it needs to minimize a joint loss. At inference stage, the organizations may have to send their individual intermediate results to the server to compute a final result. The systems of these two categories rely on cryptography and their interactions are more complex. Our FLSys focuses on Horizontal FL, with an option for extension to Vertical FL and FTL in the future. For simplicity, we will use FL to indicate Horizontal FL in the rest of our paper.

Table~\ref{table_framework_compare} shows the comparison between FLSys and other FL systems/frameworks across several features required for an efficient and effective FL system. FLSys is the only system that supports all these features, and it is also the only one that supports third-party apps and efficient mobile sensing data collection. Specifically, FLSys addresses unanswered questions on  concurrent training of multiple models for different apps and APIs for third party app developers. Furthermore, unlike all the other systems, FLSys enables models that work with data collected from the phones' sensors, which adds challenges related to efficient and effective data collection.  

Among the comparison systems, the FL work done at Google is the best known. However, despite work~\cite{bonawitz2019towards} that describes the conceptual design of a scalable FL system for mobile devices, Google has not published the implementation and evaluation of an end-to-end FL system to address the features in Table~\ref{table_framework_compare}. Recently, its TensorFlow Lite~\cite{tfl} framework started to support on-device training, but this framework does not attempt to provide any other type of system support required by FL.

Systems such as FATE~\cite{8818446} and FedVision \cite{liu2020fedvision}, introduce FL architectures based on web-services. They focus on either institutional collaboration or a target application, and they do not have any support for mobile devices. Similarly, Nvidia's FLARE~\cite{flare} is a domain-agnostic, open-source, and extensible SDK for FL, but it does not support mobile device training. 
Among the systems supporting mobile devices, Syft~\cite{pysyft} offers KotlinSyft for on-device training and provides an FL server, PyGrid, with a web-UI. However, Syft does not address scalability or provides advanced privacy preserving mechanism.
FedML ~\cite{he2020fedml} shares some goals with FLSys. However, this open source system is still under construction. In addition, FedML focuses more on software engineering aspects, rather than on system aspects such as efficient sensor data collection or scalability. 
The closest FL system to ours is Flower~\cite{beutel2021flower}, which provides a high-level FL programming library, employs TensorFlow Lite for on-device training, and evaluates scalablity with a number of embedded edge computing devices. However, this system does not focus on mobile devices and does not provide a solution to support third-party apps or mobile sensing data collection. The evaluation is conducted on embedded edge computing devices instead of real mobile devices.  Last but not least, FLSys is the only system designed to provide modular deployment. The policies, algorithms, and functions are implemented at fine granularity. The system can be deployed as interchangeable modules with serverless cloud resources, instead of an always-on server. This makes it easy to both upgrade the system and achieve cost-efficiency when scaling up.

\begin{table}[t]
\centering
\caption{Comparison of different FL frameworks (* denotes planned feature) } 
\vspace{-5pt}
\label{table_framework_compare}
\resizebox{0.49\textwidth}{!}{
\begin{tabular}{|l|l|l|l|l|l|l|l|}
\hline
                                    & TF-Lite & Syft  &  FLARE &  FATE  & FedML & Flower & FLSys \\ \hline
On-device training                  & \checkmark&\checkmark& &             &*           & \checkmark & \checkmark  \\ \hline
Scalability                         & &   &        &             &            & \checkmark & \checkmark  \\ \hline
Fault-tolerance                     & &\checkmark& &             &            & \checkmark & \checkmark  \\ \hline
Client heterogeneity                & &\checkmark &&             & \checkmark & \checkmark & \checkmark  \\ \hline
Advanced privacy preserving        & &*          &\checkmark& \checkmark  &            & \checkmark & \checkmark  \\ \hline
Concurrent third-party app support  & &      &     &             &            &            & \checkmark  \\ \hline
Efficient sensor data collection    & &       &    &             &            &            & \checkmark   \\ \hline
Modular deployment              & &         &  &\checkmark   &            &            & \checkmark   \\ \hline
\end{tabular}}
\vspace{-15pt}
\end{table}


\subsection{Coping with Heterogeneity in FL}

A well-reported issue restricting the performance of FL models is the resource and data heterogeneity among users. Resource heterogeneity arises as the on-device training performs at devices with varying computational and communication capabilities. Data heterogeneity arises because either the numbers of training samples are different, or the classes and features are non-IID ~\cite{konecny2016federated, zhao2018federated}.

To mitigate device heterogeneity, Chai et al.~\cite{chai2020tifl} divides clients into tiers based on their training time, and updates the tiers on-the-fly based on the observed training time and accuracy. However, training time is determined by both data amount and computation power. In real world, the training time has to be reported by the clients, which may introduce an additional opening for malicious users. To address device heterogeneity, FLSys adapts asynchronous communication. The server provides support for flexible client selection policies, and allows clients to self-select based on their current resources. 

Different from centralized learning, the datasets among different users may follow different distributions in FL, due to imbalanced class distributions, different user behaviors, etc. 
As a result, DL models trained in FL algorithms usually suffer from inferior performance when compared with centralized models~\cite{konecny2016federated}.
To mitigate the data heterogeneity issue, Reddi et al.~\cite{reddi2020adaptive} propose three different adaptive algorithms as aggregators for the server to aggregate client updates. 
There are studies ~\cite{DBLP:journals/corr/abs-1812-06127,sarkar2020fedfocal,zhao2018federated,9141436,10.1117/12.2519621,gao2022feddc,marfoq2022personalized,li2021fedbn,luo2021no} beyond aggregation algorithms on the server to mitigate non-IID. In FedProx~\cite{DBLP:journals/corr/abs-1812-06127}, a regularization is introduced to mitigate the gradient distortion from each device. Sarkar et al.~\cite{sarkar2020fedfocal} presented a cross-entropy loss to downweigh easy-to-classify examples and focus training on hard-to-classify examples. Verma et al.~\cite{10.1117/12.2519621} propose to estimate the global objective function by averaging different objective functions given a common region of features among users, and keep different objective functions estimated from local users' data in different regions of the feature space.  In FedDC~\cite{gao2022feddc}, the authors propose to add the penalized term, and a gradient correction term on the top of the local empirical loss term in the objective function, and each client corrects its local model parameters using the local drift variables. Data augmentation approaches have been proposed~\cite{zhao2018federated}, including a global data distribution based data augmentation~\cite{9141436}. Thanks to the modular design, FLSys can add or swap its components to support new mechanisms in FL. Our HAR-Wild and SA models use a uniform data augmentation method to achieve the best model accuracy. We also implement and show the results of different aggregators in FLSys.


\subsection{Human Activity Recognition}

Our HAR model focuses on sensing and classification of physical activities through smart phone sensors. 
Recent works show that deep learning models are effective in HAR tasks. For example, Ignatov~\cite{ignatov2018real} proposed a CNN based model to classify activities with raw 3-axis accelerometer data and statistical features computed from the data. Several works~\cite{murad2017deep,hernandez2019human,chen2016lstm} proposed LSTM-based models and achieved similar performance. 

Most research on HAR models uses centralized learning on data collected in controlled lab environments with standardized devices and controlled activities, in which the participants only focus on collecting sensor data with a usually high and fixed sampling rate frequency, i.e., 50Hz or higher.
Although there are good publicly available HAR datasets, e.g., WISDM~\cite{kwapisz2011WISDM}, UCI HAR~\cite{anguita2013ucihar}, and Opportunity~\cite{chavarriaga2013opportunity}, they are not representative for real-life situations. Different from existing works, this paper shows that HAR-Wild over FLSys performs well on data collected in the wild, which are subject to fluctuating sample rates (e.g., the sampling may be decreased temporarily to save battery power) and non-IID data distribution.

\section{FLSys Design} 
\label{sec:design}
This section presents the design of FLSys. Specifically, it describes the system requirements derived from an application-system co-design, the novel FLSys architecture that addresses these requirements, along with the four operation phases of FLSys, namely data collection and processing, privacy protection, federated training, and inference at the phones. 


\subsection{System Requirements}
\label{sec:req}

Our aim is to design and build an FL system that addresses the questions mentioned in Section~\ref{sec:intro}. We use the HAR model, detailed in Section~\ref{sec:model}, to illustrate an entire category of FL models based on mobile sensing data collected in the wild. We extract seven key requirements derived from this model and from other real-world FL applications, such as next word prediction, on-device search query suggestion~\cite{yang2018applied}, on-device robotic navigation \cite{liu2019lifelong}, on-device item ranking \cite{bonawitz2019towards}, object recognition \cite{liu2020fedvision}, sentiment analysis, etc., and utilize them to guide our FLSys design: (\emph{R1}) Effective data collection: The data collection on the phone must balance resource consumption (e.g., battery) with sampling rates required by different models; (\emph{R2}) Support for advanced privacy preserving mechanisms: Even though FL is privacy-preserving by design, there are still potential privacy issues (e.g., learn user information from the gradients)~\cite{8241854, 10.1145/3133956.3133982}. Therefore, the system must provide a plugin interface for advanced privacy protection mechanisms, such as local differential privacy; (\emph{R3}) Tolerate phone unavailability during training: Since the phones may sometimes be disconnected from the network or choose not to communicate to save battery power, the interaction between the phones and the cloud must tolerate such unavailability during federated training; (\emph{R4}) Scalability: The cloud-based FL server of our system must be able to scale to large numbers of users in terms of both computation and storage; (\emph{R5}) Model flexibility: The system must support different DL models for different application scenarios and different aggregation functions in the cloud; (\emph{R6}) Support for third-party apps: The system must provide programming support for third party apps to concurrently access different models on the phones, while efficiently managing resource consumption and contention; and (\emph{R7}) Modularity: The system shall not be heavy to deploy, and its policies, algorithms, and functions shall be designed and implemented as interchangeable modules for simple, cost-effective deployment and scalability. 
\vspace{-5pt}

\begin{figure*}[t!]
 \centering
 \vspace{-10pt}
 \subfloat[\begin{scriptsize}FLSys Architecture\end{scriptsize}]{\label{sys-arch}\includegraphics[scale=0.25]{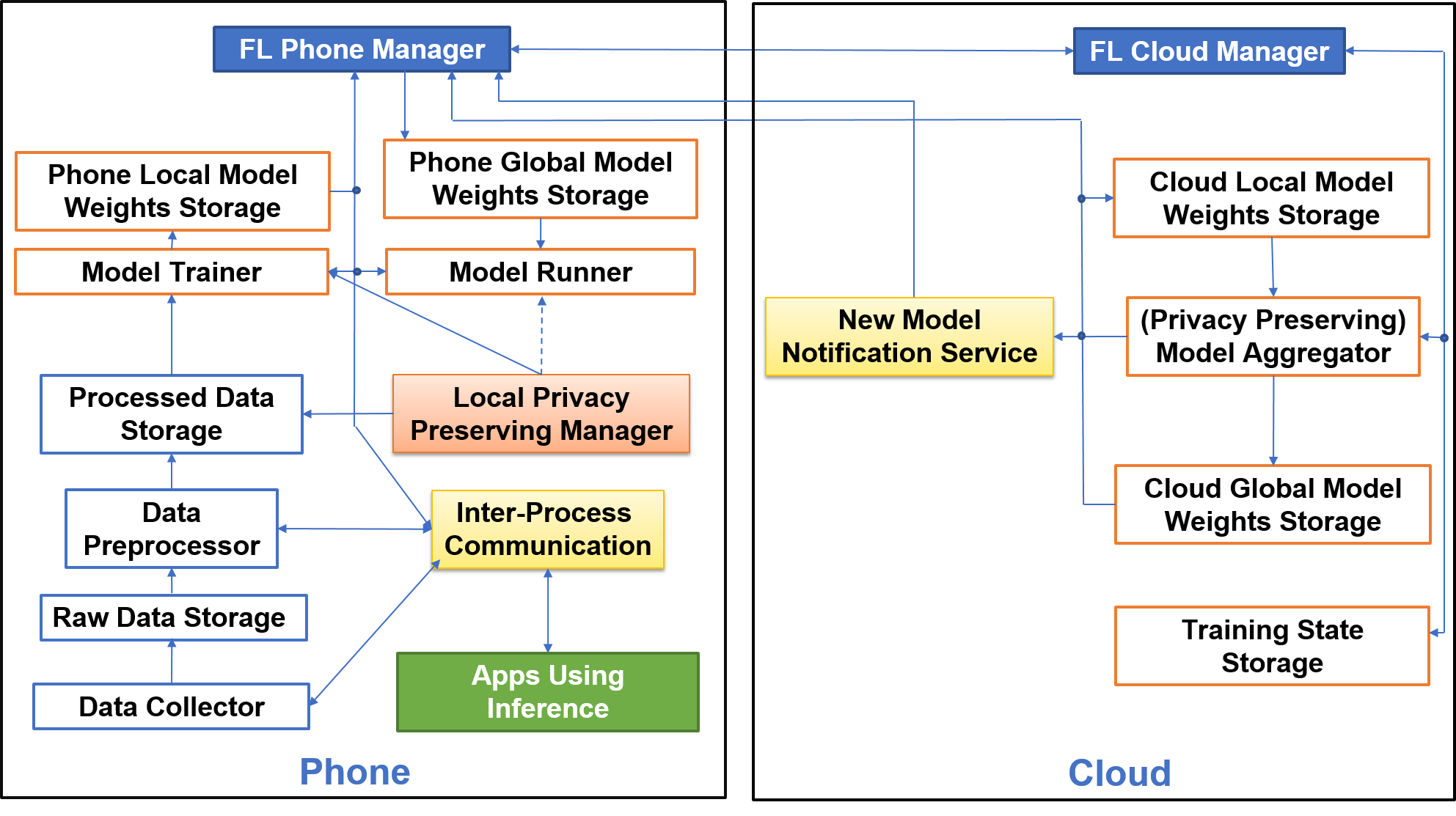}}\hfill
\subfloat[\begin{scriptsize}Asynchronous Protocol with Phone Self-Selection and Multiple Models\end{scriptsize}]{\label{FLSysAsynchronous}\includegraphics[scale=0.3375]{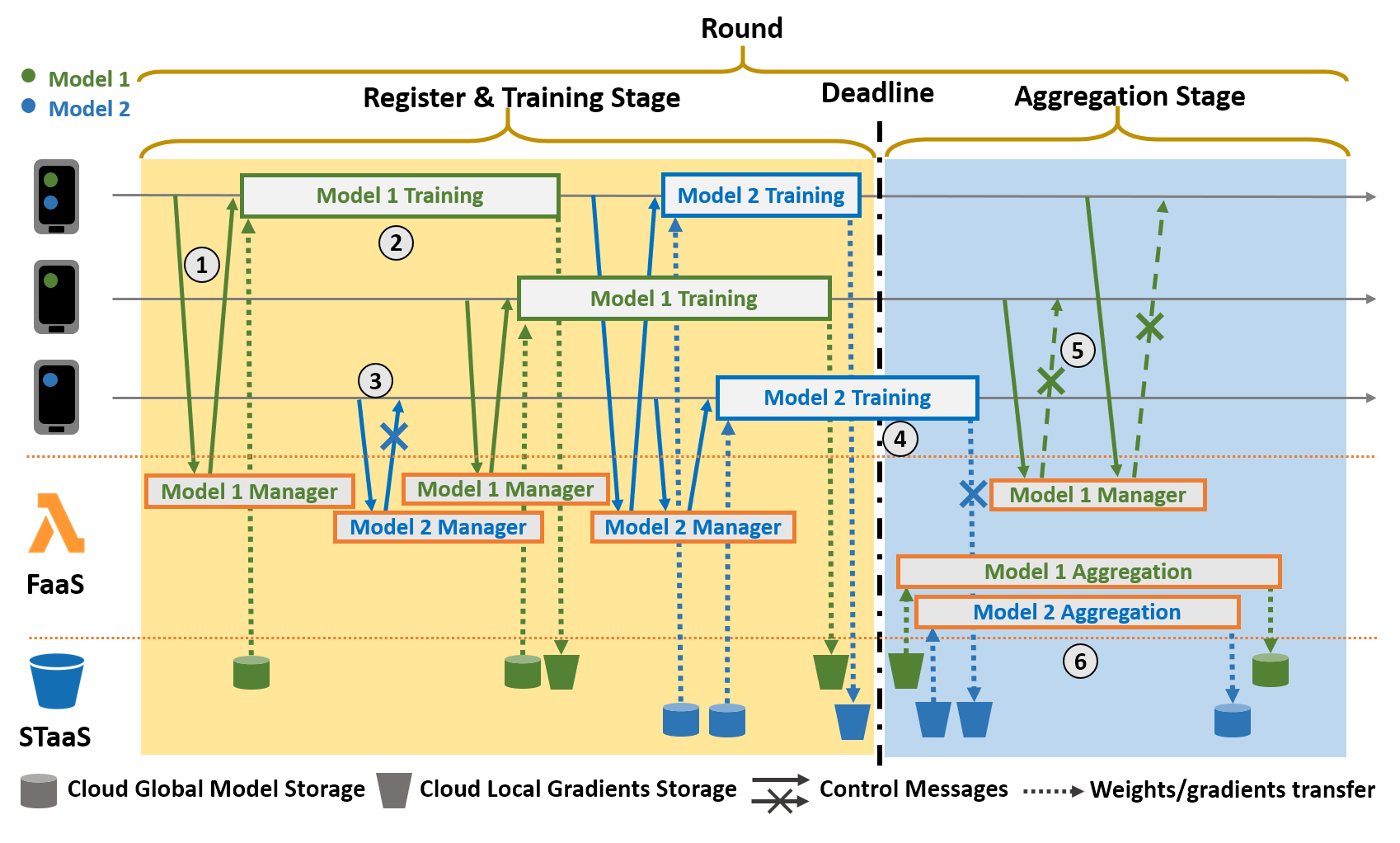}}\par 
\caption{\begin{footnotesize}FLSys Architecture and Asynchronous Protocol. Typical operations: \textcircled{1} Phone Manager of Client \#1 registers with the Cloud Manager of Model 1, which grants registration based on training settings. \textcircled{2} Phone Manager of Client \#1 fetches up-to-date global model from a designated storage, trains it with local data, and uploads local gradients to a designated storage. \textcircled{3} Phone Manager of Client \#2 tries to register, but is denied. \textcircled{4} Phone Manager of Client \#2 successfully registers at a later time, but the training misses the deadline, thus its gradients upload is denied.  \textcircled{5} Clients \#1 and \#2 try to register during server aggregation and are denied.  \textcircled{6} Each model's Aggregator loads the gradient updates, aggregates them, and saves the aggregated model.\end{footnotesize}}
\vspace{-12.5pt}
\label{sys-arch-protocol}
\end{figure*}


\subsection{FLSys Overview} 


FLSys addresses requirements $R1-R7$ synergistically in a novel system architecture. For some requirements, we propose novel solutions, as no current FL system addresses them, while for others we customize existing solutions for our needs in order to provide a complete design and implementation. Figure~\ref{sys-arch} shows the system architecture, and Figure \ref{FLSysAsynchronous} shows the overall process of one training round. These figures emphasize five novel contributions made in FLSys, compared with existing FL systems: \textbf{(1)} FLSys allows the phones to self-select for training when they have enough data and resources; \textbf{(2)} FLSys has an asynchronous design (Figure~\ref{FLSysAsynchronous}), in which the server in the cloud tolerates client failures/disconnections and allows clients to join training at any time. \textbf{(3)} FLSys supports multiple DL models that can be used concurrently by multiple apps; each phone trains and uses only the models for which it has subscribed; \textbf{(4)} FLSys acts as a ``central hub'' on the phone to manage the training, updating, and access control of FL models used by different apps; and \textbf{(5)} FLSys allows apps/models to use different privacy mechanisms that trade model accuracy for privacy guarantees.

These features balance model utility with mobile device constraints and privacy, and can help create an ecosystem of FL models and associated apps. FLSys allows different developers to build FL models/apps and provides a simple way for users to take advantage of these apps, as it offers a unifying system for the development and deployment of FL models and apps that use these models. FLSys acts as common middleware layer for all these apps and models. The users just need to download/install the apps, and FLSys will take care of downloading/installing the FL models used by the apps, will perform FL training as needed, and will run FL inference on behalf of the apps.

\subsection{System Architecture} 

The architecture (Figure~\ref{sys-arch}) has two main components: \emph{(1)} \emph{FL Phone Manager}, which coordinates the FL activities on the phone; and \emph{(2)} \emph{FL Cloud Manager}, which coordinates the FL activities in the cloud. These two components work together to support the four phases of the FL operation: data collection and preprocessing, privacy protection, model training and aggregation, and mobile apps using inference. In the following, we describe each phase and explain how the system architecture satisfies the seven system requirements.

\textbf{Data Collection and Preprocessing.}
The FL Phone Manager controls the data collection using one or multiple \emph{Data Collectors}. A basic Data Collector is tasked with collecting data from one sensor at a given sampling rate. Such basic Data Collectors could be embedded in more complex ones to collect different types of data at the same time. It is important to have one app that coordinates data collection because having multiple apps collecting overlapping sets of data multiple times is inefficient. 
Having the FL Phone Manager to coordinate the data collection also simplifies sensor access control.

To satisfy requirement \emph{R1}, FLSys supports on-demand configuration of sensor types, sampling rates, and the period to flush data from memory to storage. Each model informs the FL Phone Manager of the type of data and sampling rate it needs. In this way, the FL Phone Manager knows which Data Collectors to invoke and which sampling rates are needed.
The FL Phone Manager balances sensing accuracy (i.e., high sampling rate) with resource consumption. 

To regulate and keep such balance aligned with the user experience, the FLSys has three features: (1) include several built-in sampling rate settings, with empirical values from our experience; and (2) collect key statistics of the data collection (e.g., CPU time consumed, battery life impact, etc.) and show them to the user, upon request; and (3) provide global level controls for the user to adjust the data collection behaviors, should the user feel that their experience is impacted by data collection.

The Data Collectors store the sensed data in the \emph{Raw Data Storage} and inform the FL Phone Manager each time new data is added to the Raw Data Storage. For efficiency, the Data Collectors can buffer a certain amount of sensed data in memory before committing it to the storage. The FL Phone Manager can dynamically reconfigure the data flushing period that defines when the data is written to storage. Data Collectors set this data flushing period. 
Some models may use the raw data directly, while others may require additional processing. The FL Phone Manager decides when to invoke the model-specific \emph{Data Processors}, which will store the data in the \emph{Processed Data Storage}. This is a matter of policy and can be done any time new data is available in the Raw Storage Data or at a regular interval. The only constraint is to have all the data preprocessed before a new local model training operation. 

To deal with the problem of non-IID data distribution, described in Section~\ref{sec:related}, the \emph{Data Preprocessor} can augment the data collected locally on the device with data received from the cloud. The augmentation dataset is model-specific and mitigates the distortion the data classes by providing data samples for classes with not enough data. When the users join FLSys for a new model, their phones receive an augmentation dataset from the \emph{FL Cloud Manager} for models that use data augmentation techniques.

\new{
\textbf{Privacy Threat Model and Protection}. To satisfy requirement \emph{R2}, the \emph{Local Privacy Preserving Manager} delivers advanced privacy protection mechanisms on the phone component of FLSys. It is designed to work with different privacy mechanisms, which are available on a per-model basis. 

\textit{Threat Model.} In this paper, we focus on defending against privacy inference attacks from an honest-but-curious server, which can attempt to infer clients' local training data. Note that the server knows the identity of the clients to coordinate the FL training. The server may try to extract the clients' local data or infer membership information of specific clients' training data samples by using training data extraction attacks \cite{carlini2020extracting} and membership inference attacks \cite{DBLP:conf/sp/NasrSH19,10.1145/3359789.3359824, DBLP:journals/corr/HitajAP17}, respectively, via observing the clients' local gradients. Third-party apps on the phones, which may or may not use FLSys, may act maliciously by trying to access the model data or performing inference attacks, etc. There are many OS-based, programming language-based, and networking-based approaches that can prevent or alleviate these issues. All these solutions can be applied outside of FLSys.

\textit{Defenses.} An effective way to protect clients' local training data against an honest-but-curious server is to use local differential privacy (LDP)~\cite{kim2021federated}, specifically to preserve $\epsilon$-LDP in FL \cite{sun2020ldp}. LDP provides rigorous privacy protection, without computational overhead, compared with other techniques such as secure multi-party computation \cite{tian2022flvoogd} and homomorphic encryption \cite{wagh2021dp}. Meanwhile, anonymizers (shuffler \cite{liu2020flame}, faking source IP, VPN, Proxy, mixnets, etc. \cite{sun2020ldp}) could be compromised or could collude with the server to extract sensitive information from local gradients \cite{erlingsson2019amplification}. This introduces extra privacy risks for the clients' local training data. In addition, it is challenging for dimension reduction-based privacy-preserving techniques to achieve good utility under rigorous privacy guarantees with complex models and tasks \cite{liu2020fedsel}.

Therefore, our system supports existing LDP-preserving approaches in FL, which are currently the most suitable solutions.
Existing LDP-preserving approaches in FL can be divided into two categories: \textbf{(1)} Clients add noise to local gradients to protect the values of the local gradients \cite{ malekzadeh2021dopamine}; and \textbf{(2)} Clients add noise to each training sample to protect the value of each training sample \cite{lyu2020towards}, and then they use these perturbed samples to derive local gradients. For both approaches, the clients send the LDP-preserved local gradients to the server for model updates.
Our system further supports \emph{User-level DP} (\emph{User-DP}) \cite{mcmahan-userDP} to protect the membership information of clients against inference attacks.

These supported mechanisms \cite{mcmahan-userDP,BitRand} use DP to provide different levels of privacy protection. Within a DP budget allocated to a given privacy mechanism, the global model converges without an undue cost of model utility. In \emph{User-DP}, the aggregated gradients at the FL Cloud Manager are perturbed to protect clients' participation (membership) information in training the global model (Definition 2, Appendix A).
In \emph{LDP}, every training sample is perturbed under LDP ensuring that the legitimate value of the training sample is protected against being inferred by the server through observing local gradients (Definition 3, Appendix A). As these two mechanisms illustrate, the \emph{Model Aggregator} in the cloud may (e.g., User-DP) or may not (e.g., LDP) apply privacy preserving mechanisms. Generally speaking, privacy mechanisms in FLSys are handled by the \emph{Local Privacy Preserving Manager} on the phones, with potential collaboration from the \emph{Model Aggregator} in the cloud.
}

\textbf{Federated Training.} 
To satisfy requirement \emph{R3}, we make two design decisions. First, FLSys allows the phones to self-select for training when they have enough data and resources. This is different from traditional FL architectures~\cite{bonawitz2019towards}, where the server selects the phones to participate in training, which may not be available or may not have enough data or resources for training. Second, in FLSys, the communication between the phones and the cloud is asynchronous to cope with phone disconnections. The software at the cloud side is designed to tolerate missing messages from the phones. Overall, FLSys reduces communication overhead and increases client utility, at the expense of less control in the client sampling process, compared to~\cite{bonawitz2019towards}.

In order to use a given model on the phone, the FL Phone Manager first registers the phone with the \emph{FL Cloud Manager}. If the phone model and mobile OS are known to work with the model, the FL Cloud Manager registers the phone with the \emph{New Model Notification Service}, which works as a Publish-Subscribe cloud service, and returns the subscription to the phone. This subscription allows the phone to receive asynchronous notifications when a new global model is available for download. The FL Phone Manager downloads the model at a time determined based on the model usage frequency and power settings.

The training for each model is done in rounds. The FL Cloud Manager decides the duration of a round, based on preferences associated with each model. For example, the server may start a new aggregation (i.e., by invoking the \emph{Model Aggregator} for a certain model) when a given time interval has passed or when a certain number of local training updates have been received from the phones. The FL Phone Manager decides when to participate in training. This decision is done based on local policies that attempt to balance inference accuracy, the amount of input data for training, and the resources consumed during training. The intention to participate in training for a given model is conveyed by a message sent to the FL Cloud Manager. Based on the model preferences (e.g., amount of data, and the number of users in a training round), the server may decide to ask the phone to train for the model and to provide the FL Phone Manager with a URL to upload the results in the \emph{Cloud Local Gradients Storage}. If there is a deadline for participation in the round, the FL Cloud Manager lets the FL Phone Manager know about it.

The FL Phone Manager invokes the \emph{Model Trainer} for the given model and passes as parameter the location of the data in the Processed Data Storage. After the training is done, the Model Trainer stores the newly computed gradients in the \emph{Phone Local Gradients Storage}. The FL Phone Manager decides when to upload these gradients to the Cloud Local Gradients Storage. The FL Cloud Manager will invoke the Model Aggregator for the model when the duration for the round expires or when enough updates have been uploaded. The Model Aggregator reads the updates from the Cloud Local Gradients Storage, computes the aggregated weights, and stores them in the \emph{Cloud Global Model Weights Storage}. The intermediate training state is stored in the \emph{Training State Storage} to provide lower I/O latency compared with the other types of cloud storage in our design. This is because FLSys needs frequent access to these data during training. Then, the Model Aggregator sends a notification via the New Model Notification Service to let the phones know that a new model version is available.

The cloud-side system satisfies requirement \emph{R4}, as it can scale to large numbers of users due to its modular design that decouples computation, communication, storage, and notification services. The cloud elasticity features of each service allow different services to scale up or down according to the workload. 

As we observe from the architecture, each model is managed individually by FLSys, and multiple models can co-exist both at the phones and the cloud. In the cloud, different models use independent cloud resources, which can be scaled independently. On the phone, independent model trainers and inference runners are responsible for different applications. The cloud contains all the models in the system, while each phone contains only the models for which it has subscribed. This modular design allows our system to satisfy requirement \emph{R5}.

\textbf{Mobile Apps Using Inference.}
We decouple mobile apps that need inference on the phones from the models that provide the inference. This allows an app to use multiple models, while the same model can be used by multiple apps. FLSys provides an API and a library that can be used by third-party app developers to perform inference using DL models on the phone. In this way, the system architecture satisfies requirement ~\emph{R6}. When an app needs inference from a model, it sends a request to the FL Phone Manager using one of the OS IPC mechanisms. 
The FL Phone Manager then generates the input for the inference from the data stored in the Processed Data Storage or the Raw Data Storage, and then invokes the \emph{Model Runner} with this input. The Model Runner sends the result to the App using IPC. When possible, the FL Phone Manager re-uses preprocessed data to reduce resource consumption or performs one inference for several applications that invoke the same model concurrently.

\textbf{Model Concurrency.}
Given the design of FLSys, both the FL Phone Manager and the FL Cloud Manager are able to handle multiple models concurrently. However, the meanings of concurrency are slightly different for each side. FL Cloud Manager needs to handle the aggregation of all models that are registered with it. Also there is the need to communicate to a potentially large number of clients for each model at the same time. FLSys handles this concurrency through services provided by the underlying cloud platform, which support concurrency by design. FLSys just needs to orchestrate the invocation of these services.
The FL Phone Manager needs to handle concurrent training and inference. Our preliminary experiments on smart phones show parallel training of multiple models is very slow due to resource contention. It also affects the user experience on the phones. Therefore, we decided to train models sequentially. The FL Phone Manager can request to participate in training rounds for multiple models concurrently, but it locally decides a sequential order in which to train these models, based on parameters such as frequency of model usage by apps, the training round deadlines, and historical training latency for each model. Finally, the inference requests from the apps are executed as soon as they are received to maintain good user experience.

\textbf{System Modularity.}
FLSys components are designed and implemented at fine granularity as interchangeable modules for different policies and algorithms to satisfy requirement \emph{R7}. This design makes it easy to deploy different data collection modules, DP-based privacy preserving mechanisms, model trainers at the clients (with different optimizers or loss functions), and aggregation functions at the server. 
 Furthermore, new models can be added on-demand, based on the apps that need them. This modular design can be readily deployed in a serverless manner in the cloud, which leads to improved scalability (i.e., scale up only the components that are overloaded) and cost-efficiency (i.e., no need to run always-on servers).

\section{Prototype Implementation} 
\label{sec:implementation}
We implemented an end-to-end FLSys prototype in Android and AWS cloud, which have been chosen because they are the market leaders for mobile OSs and cloud platforms, respectively. However, the FLSys design is general and it can be implemented in other mobile OSs and cloud platforms. The prototype implements all of the components described in the system architecture (Figure~\ref{sys-arch}). This section reviews the implementation technologies, the reasons for selecting them, and then focuses on the Android implementation and the AWS implementation of FLSys. 



\subsection{Implementation Technologies} 


\textbf{Deep Learning Framework.}
We chose Deep Learning for Java (DL4J) as the underlying framework for the on-device DL-related operations (i.e., training and model execution) because it was the only mature framework that supported model training on Android devices until very recently, when TensorFlow Lite~\cite{tfl} and KotlinSyft~\cite{pysyft} became available for on-device training. While the Model Aggregator in the cloud could be implemented using other DL technologies, for consistency, we implement it in DL4J as well. The models are stored as zipped JSON and bin files in folders on the phone and in AWS S3 buckets in the cloud.

\textbf{On-device Communication.}
For IPC among Android apps/services, we use Android Bound Service and Android Intent. A bound service can efficiently serve another application component because it does not run in the background indefinitely. 
Through IPC, the FL Phone Manager can provide third-party apps with an interface to request inference results without revealing the model or the data. Furthermore, it can communicate with the Data Collector.

\textbf{Cloud Platform and Services.}
We opt to utilize the Function-as-a-service (FaaS) architecture for our cloud computation. The core cloud components of FLSys are implemented and deployed as AWS Lambda functions~\cite{lambda}. We decided to choose FaaS for our implementation for five reasons. First, it matches our asynchronous, event-based design, as Lambda functions are triggered by events. Second, it provides fine-grained scalability at the function level; therefore leading to less resource consumption in the cloud. Furthermore, computation and storage are scaled automatically and independently by the cloud platform. Third, unlike other cloud platforms, it does not require running virtual machines when no computation is necessary; this saves additional resources and reduces cost. Fourth, FaaS simplifies the development and deployment of our prototype because it does not require software installation, system configuration, etc. Fifth, different functions can be implemented in different programming languages making the implementation even more flexible.

Lambda functions are triggered in different ways in our prototype. We use the AWS API Gateway to define and deploy HTTP and REST APIs. For instance, we create a REST API to relay clients' requests to participate in the FL training to the Lambda function that handles these requests. We also use the AWS EventBridge to define rules to trigger and filter events for Lambda functions. 

FLSys uses a number of cloud services for storage, authentication, and publish-subscribe communication. For model storage, validation datasets, and FL Cloud Manager configuration files, we use AWS S3, which offers a reliable and cost-effective solution for data accessed infrequently. More importantly, AWS S3 buckets can be accessed directly by phones, which simplifies the asynchronous communication in FLSys. To authenticate clients and allow them to upload and download models from the AWS S3, FLSys uses Identity Pool in AWS Cognito. To store data that is accessed frequently, such as training round states and model states, we use AWS DynamoDB, a reliable NoSQL database. AWS SNS is utilized in conjunction with the Google FCM to notify clients when newly trained models are ready. The use of a Google Cloud service in our AWS implementation was necessary in order to push notifications directly to apps on the phones when a new global model is ready in the cloud. 


\subsection{Phone Implementation} 

The phone implementation (left-side of Figure~\ref{sys-arch}) consists of three apps: a FL Phone Manager, a HAR Data Collector, and a Testing App used to test model inference. 

\textbf{Data Collector.}
\label{sec:dc}
We implemented a HAR Data Collector app designed for long-term and battery efficient data collection. This Data Collector was implemented as an app that can be used independent of FLSys, but for better efficiency, the Data Collectors can be implemented as modules of the FL Phone Manager. To that end, sensor values are not collected at an enforced fixed high frequency, but are instead collected independently through Android listeners whose actual frequency is variable, determined by the underlying OS. This is appropriate for data collection in the wild. In our experience, this tends to be much friendlier to the performance and battery life of the user devices, lowering the risk that a user abandons FLSys prematurely due to concerns about how it is affecting their device resources. Furthermore, users are given the option to pause or stop data collection of all or a subset of sensors in case they have resource consumption or privacy concerns. For simplicity, the raw data and the processed data are stored as files. 

\textbf{FL Phone Manager.} \label{fl_phone_manager}
The FL Phone Manager app decides to initiate an on-device training round based on evaluating a Ready To Config policy (RTCp). We implemented a simple policy to check if the phone is charging and is connected to the network before declaring its availability for training. 
If yes, it sends a Ready To Config message (RTCm) to the FL Cloud Manager. 
RTCm is implemented as an HTTP request with JSON payload and is sent to a REST API URL in AWS. The FL Cloud Manger either accepts or denies the phone's participation in this training round, based on a simple Accept/Deny for Training policy (A/DFTp) that checks the phone model and client identity.

The phone is accepted for a round of training when it receives an Accept For Training message (AFTm). AFTm contains the information of the AWS S3 locations from where to download the latest global model weights and where to upload the local gradients. The message also contains the deadline for this training round's completion. 
The FL Phone Manager evaluates a Start To Train policy (STTp) based on the available device resources and the round's deadline to determine whether to actually perform the on-device training for this round or not.

The FL Phone Manager will create the corresponding Model Trainer if it decides to train. The Model Trainer is implemented with Android native \emph{AsyncTask} class to ensure the trainer is not terminated by Android, even when the app is idle. \emph{AsyncTask} also enables multiple trainers to train in the background. Once the training is complete, the Model Trainer uploads the local gradients to the corresponding AWS S3 location. 

Model inference is implemented as a background service with Android Interface Definition Language (AIDL), and it gets inference requests from third-party apps. When such a request is received, the FL Phone Manager uses the current sensor data from the Data Collector as input for the model, runs the inference, and responds to the third-party apps with the inference results. 

\textbf{Testing App.} 
We implemented a simple testing App to test model inference. The App uses \emph{AidlConnection} to interface with the FL Phone Manager. Let us note that the App itself does not access any data or model.


\subsection{Cloud Implementation} 


The cloud implementation (right-side of Figure~\ref{sys-arch}) consists of two main components: FL Cloud Manager and Model Aggregator.

\textbf{FL Cloud Manager.}
The FL Cloud Manager is implemented as a series of Lambda Functions (FaaS service in AWS). 
When starting a training round, it reads a configuration file and determines the deadline for the round (i.e., the time when the round must finish). During the period between the start time and the deadline, the FL Cloud Manager accepts or denies clients' requests for training (RTCm). When the deadline is reached, the FL Cloud Manager executes the Model Aggregator according to the Start for Aggregation policy (SFAp). The current policy checks if enough clients have submitted their local gradients in the AWS S3 (a configurable parameter). Then, the Lambda function implementing the FL CLoud Manager schedules an event for itself to perform the next training round and terminate. The training process stops when the pre-defined number of rounds is achieved, or the desired performance (model accuracy) is achieved, if the model developers provided a validation dataset.

\textbf{Model Aggregator.}
For implementation simplicity, the Model Aggregator uses the federated average technique~\cite{mcmahan2017communication}, with the assumption that each client contributes equally to the global model in each training round.
When it is invoked, it loads the uploaded local gradients, and aggregates their gradients to the global model of this round. Once the global model is updated, the Model Aggregator invokes AWS SNS to notify clients that they can download the newly aggregated model. Note that the Model Aggregator is called dynamically through reflection, such that different aggregation functions can be dynamically swapped. 

\subsection{Asynchronous Federate Averaging Implementation}

Algorithm~\ref{alg_async_fedavg} shows the pseudo-code of our asynchronous federated averaging process. The algorithm consist of three procedures, which execute asynchronously. ``ClientLoop'' (lines 1-12) runs at clients and executes a round of training (lines 7-12), if the phone self-selects for training and the cloud accepts it (lines 1-6). ``ServerRTCmHandler'' (lines 13-17) is a part of the FL Cloud Manager and decides whether a phone is accepted for training. ``ServerLoop'' (lines 18-40) also runs at the FL Cloud Manager. It performs the aggregation of local gradients and controls the progression of training. The clients participating in a training round must submit their local gradients before the deadline for the round expires. When the deadline comes, the procedure first evaluates the Start for Aggregation policy, which checks whether there are enough local gradient updates in order to preform aggregation. If yes, the aggregation is preformed (line 24-26); if not, this round is aborted, but the uploaded gradient updates will be carried to the next round. After aggregation, the procedure may check against pre-defined conditions to decide whether this aggregation outcome should be accepted or not (lines 27-30). Finally, the procedure checks if a new round should be started by evaluating the Start New Round policy. If a new round is to be started, a new deadline will be set (lines 33-36). Otherwise, the procedure terminates.

\begin{algorithm}[h]
    \scriptsize
	\caption{AsyncFedAveraging~\label{alg_async_fedavg}} 
	\begin{algorithmic}[1]
	    \Procedure{ClientLoop}{}
	        \While{true}
	            \State $readyToConfig \gets$ \Call{evaluateReadyToConfigPolicy}{$powerState$, $wifiState$,...}
	            \If{$readyToConfig$}
	                \State $response \gets$ \Call{sendRTCm}{ }
	                \If{$response$ == ``AFT''}
	                    \State $\mathcal{B} \gets$ \Call{sampling}{$\mathcal{D_{L}}$}
	                    \State $\theta_{l} \gets \theta^t$
	                    \For{batch $b \in \mathcal{B}$}
	                        \State $\theta_{l} \gets \theta_{l}-\eta \nabla \mathcal{L}(\theta_{l};b)$
	                    \EndFor
	                    \State $\Delta_{l} \gets \theta_{l} - \theta^t$
	                    \State \Call{uploadClientGradients}{$\Delta_{l}$}
	               \EndIf
	           \EndIf
	        \EndWhile
	    \EndProcedure
	    \item[]
	   \vspace{-10pt}
	    \Procedure{ServerRTCmHandler}{$RTCm$}
	        \If{\Call{evaluateAcceptForTrainingPolicy}{$RTCm$}}
	            \State \Call{returnResponse}{``AFT''}
	        \Else
	            \State \Call{returnResponse}{``DFT''}
	        \EndIf
	    \EndProcedure
	    \item[]
	    \vspace{-10pt}
	    \Procedure{ServerLoop}{}
	        \State $deadlineTriggered \gets false$
	        \State \Call{setupDeadline}{ } ($deadlineTriggered \gets true$ when triggered)
	        \While{true}
	            \If{$deadlineTriggered$}
	                \If{\Call{evaluateStartForAggregationPolicy}{ }}
	                    \State $\{\Delta_{1},...\Delta_{k}\} \gets$ \Call{loadClientGradients}{ }
	                    \State $\Delta^{t} = (\sum_{k}\Delta_{k}) / k$
	                    \State $\theta^{t+1} \gets \theta^t + \gamma \Delta^t$
	                    \If{\Call{isRoundAcceptable}{ }}
	                        \State \Call{acceptRound}{$\theta^{t+1}$}
	                    \Else
	                        \State \Call{abortRound}{ }
	                    \EndIf
	                \Else
	                    \State \Call{abortRound}{ }
	                \EndIf
	                \If{\Call{evaluateStartNewRoundPolicy}{ }} 
	                    \State \Call{startNewRound}{ }
	                    \State $deadlineTriggered \gets false$
	                    \State \Call{setupDeadline}{ }
	                \Else
	                    \State \Call{stopTraining}{ }
	                \EndIf
	            \Else
	                \State \Call{wait}{ }
	            \EndIf
	        \EndWhile
	    \EndProcedure
	\end{algorithmic} 
\end{algorithm} 

\subsection{FLSys Setup Workflow} 

By design, FLSys acts as a service provider that handles multiple FL models with minimum input from the users. The setup procedures for FLSys are divided into two stages. The first stage involves the FL Cloud Manager and the app developers, without user involvement. The second stage involves the FL Phone Manager and the mobile apps that use FL models, and it requires minimum user involvement. The FL Cloud Manager is deployed before the first stage, and the FL Phone Manager should be installed on the user's device before the second stage. To illustrate these stages, let us briefly explain the setup workflow using the HAR app as an example.

In the first stage, the developers of the HAR app need to register the model with the FL Cloud Manager. The app developers need to provide the FL model to be trained and the training plan (e.g., training frequency, number of rounds, number of participants in a round, etc.) to register the app. The model can be developed by the app developers or by a third party. After registration, a unique key for the authentication between the app and the FL Phone Manager in the second stage will be provided. 

The second stage is typically triggered during the installation process of the HAR app on the user's device. The app will communicate with the FL Phone Manager and authenticate itself using the aforementioned unique key. Once the app is successfully authenticated, the FL Phone Manager will perform a series of operations and eventually become ready to serve the FL model for the app. These operations include: (1) Register the phone with the FL Cloud Manager; (2) Set up communication channels with the app; (3) If the model does not exist on the phone, download the model specified by the app and the training plan from the FL Cloud Manager; If the model already exists on the phone, establish the connection between the app and that model; and (4) Set up the local training schedule and notify the user.
After the second stage, the FL model that the HAR app needs is installed on the phone, ready for inference and training. The training plan can be adjusted by the developers through the FL Cloud Manager. User-experience related parameters can be adjusted by the user through the FL Phone Manager.

\section{HAR-Wild: Data, Model, and Training} 
\label{sec:model}

We co-designed FLSys with a HAR model, which was used to extract the main requirements for FLSys and, then, to demonstrate the efficiency and effectiveness of FLSys. To show that FLSys works with different concurrent models, we also implemented and evaluated a sentiment analysis (SA) model, as described in Section~\ref{sec:results}. In this section, we describe the HAR dataset, our HAR-Wild model, and its training algorithm using data augmentation to deal with non-IID data in the wild.

\subsection{Data Collection} 

Although there are good HAR datasets publicly available, e.g., WISDM~\cite{kwapisz2011WISDM}, UCI HAR~\cite{anguita2013ucihar}, they are not representative for real-life situations because they were collected in rigorously controlled environments on standardized devices and controlled activities, in which the participants only focused on collecting sensor data with a usually high and fixed sampling rate frequency, i.e., 50Hz or higher. 
Thus, given our goal to test FLSys with data collected in the wild, we have used our Data Collector, described in Section~\ref{sec:dc}, to collect data from 116 users at two universities.

The data collection was approved by the IRBs at both universities. 
The students collected data for four months. Each user provided accelerometer data and labels of their activities on their personal Android phones. We provided labels in five categories for participants to choose form: ``Walking,'' ``Sitting,'' ``In Car,'' ``Cycling,'' and ``Running''.
The phones were naturally heterogeneous, and the daily-life activities were not constrained by our experiments.

Therefore, we collected a novel HAR dataset in the wild that is different from the existing datasets in the following three aspects: \textbf{(1)} The sensors' sampling rates vary from time to time and from user to user, due to battery constrains, device heterogeneity, and usage differences; \textbf{(2)} The same basic activity will generate different signals since different users will have different habits of carrying smart phones; \textbf{(3)} Label distributions are not just biased, but vary significantly among users.

\subsection{Data Preprocessing} 
Our data processing consists of the following steps:
\textbf{(1)} Any duplicated data points (e.g., data points that have the same timestamp) are merged by taking the average of their sensor values;
\textbf{(2)} Using 300 milliseconds as the threshold, continuous data sessions are identified and separated by breaking up the data sequences at any gap that is larger than the threshold;
\textbf{(3)} Data sessions that have unstable or unsuitable sampling rates are filtered out. We only keep the data sessions that have a stable sampling rate of 5Hz, 10Hz, 20Hz, or 50Hz; 
\textbf{(4)} Data sessions are also filtered with the following two criteria to ensure good quality: (a) The first 10 seconds and the last 10 seconds of each data session are trimmed, based on our observations of the user behavior and data. The first 10 seconds allow the users enough time to completely change from one activity to another, without affecting the label annotation. The last 10 seconds allow the users enough time to finish their activity and label annotation. Without giving the users enough time to begin and end their activities, the labeled data will be in noisy. As future work, an automatic solution in FLSys may be able to adapt dynamically such cut-off points for data across different models and types of data.
(b) Any data session longer than 30 minutes is trimmed down to 30 minutes, in order to mitigate the potential inaccurate labels due to users’ negligence (forgot to turn off labeling); and 
\textbf{(5)} We sample data segments at the size of 100 data points with sliding windows. Different overlapping percentages were used for different classes and different sampling rates. The majority classes have 25\% overlapping to reduce the number of data segments, while the minority classes have up to 90\% overlapping to increase the available data segments. The same principle is applied to sessions with different sampling rates. 
We sample 15\% of data for testing, while the rest are used for training. Details are shown in Section~\ref{sec:results}.

\textbf{Data Normalization.} In our models, the accelerometer data is normalized as $x \in [-1, 1]^3$ to achieve better model utility. 
We compute the mean and variance of each axis (i.e., $X$, $Y$, and $Z$) using only training data to avoid information leakage from the training phase to the testing phase. Then, both training and testing data are normalized with z-score, based on the mean and variance computed from training data. 
Based on these results, we choose to clip the values in between $[min, max] = [-2, 2]$ for each axis, which covers at least 90\% of possible data values.
 Finally, all values are linearly scaled to $[-1, 1]$ to finish the normalization process:
\begin{equation}
x = 2 \times [\frac{x - min}{max - min} - 1/2]
\end{equation}

\begin{figure}[t]
  \centering
  \includegraphics[width=0.99\linewidth]{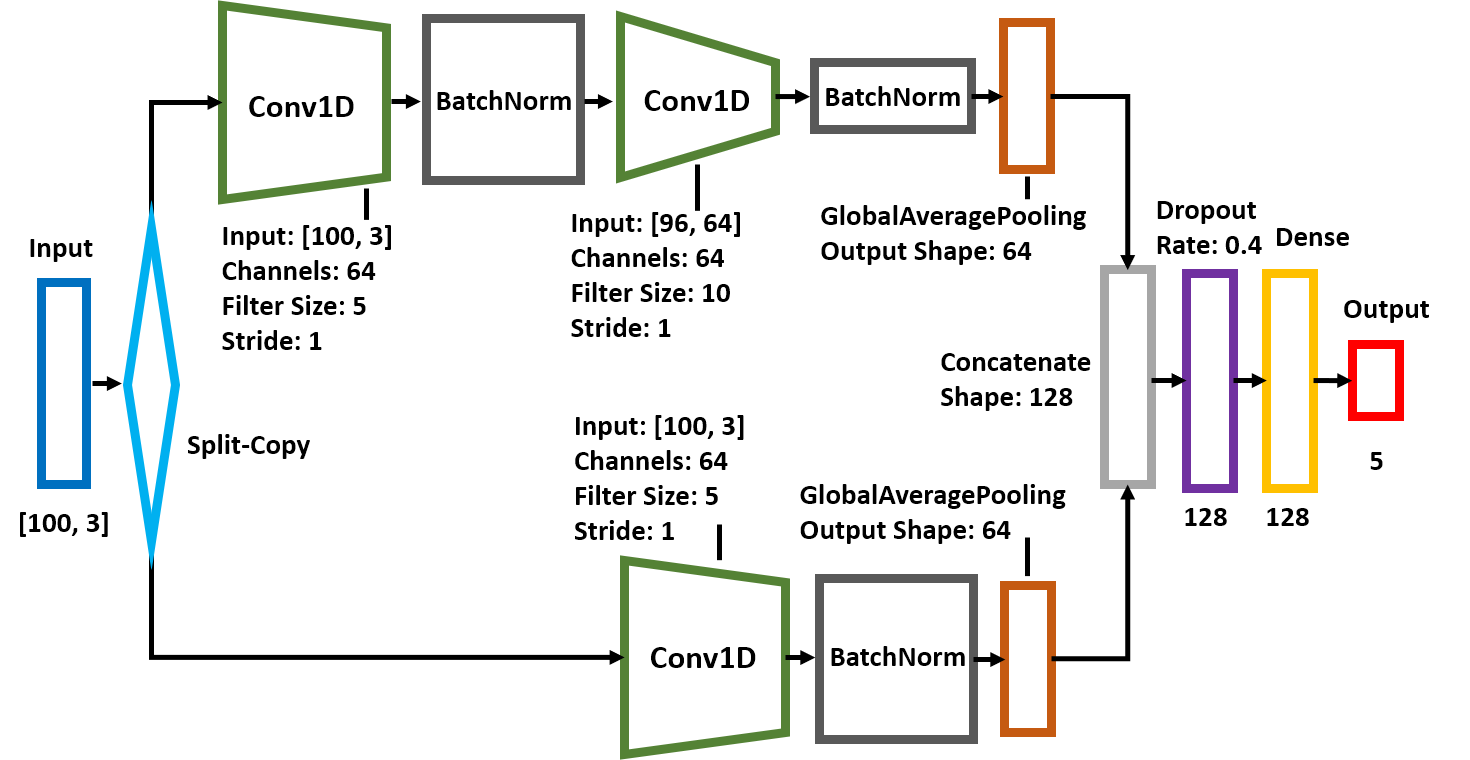}
  \vspace{-5pt}
  \caption{HAR-Wild Model Architecture} 
  \label{arch}
  \vspace{-0.15in}
\end{figure}
\subsection{Model Design} 
The design of our HAR-Wild model has two requirements: low computational complexity and small memory footprint. Satisfying these requirements ensures the model can work efficiently on resource-constrained phones. Figure~\ref{arch} shows our model architecture. For low computation complexity, HAR-Wild is based on CNN (instead of RNN, e.g., LSTM) and tailored to work well on mobile devices. In addition, instead of using data from multiple sensors, HAR-Wild can achieve comparable results with several baseline approaches by using only accelerometer data, which makes the training faster. 

The accelerometer data are processed into data segments of shape $[3, 100]$, indicating 100 data points of 3 axis: X, Y, and Z. We leverage the recipe of ResNet model \cite{he2016deep} into a small-size model, by using the processed accelerometer data as input of (1) a sequence of a 1D-CNN - a Batch Norm - a 1D-CNN - a Batch Norm - a Flatten layer, and (2) a sequence of a 1D-CNN - a Batch Norm - a Flatten layer. The two flatten layers are concatenated before feeding them into a sequence of a Drop Out layer - a Dense layer - and an Output layer. By doing so, HAR-Wild can memorize and transfer the low level latent features learned from the very first 1D-CNN, directly derived from the input data, to the output layer for better classification. We use Global Average Pooling~\cite{lin2013network} given its small memory footprint, instead of the popular Local Max/Average Pooling~\cite{goodfellow2016deep}. In addition to being appropriate for resource-constrained phones, a small-size model such as HAR-Wild is expected to perform better on data collected in the wild, since the data will likely have more distribution drift, increasing the chance of model overfitting on large-size models.

\begin{figure}[t]
  \centering
  \includegraphics[width=0.85\linewidth]{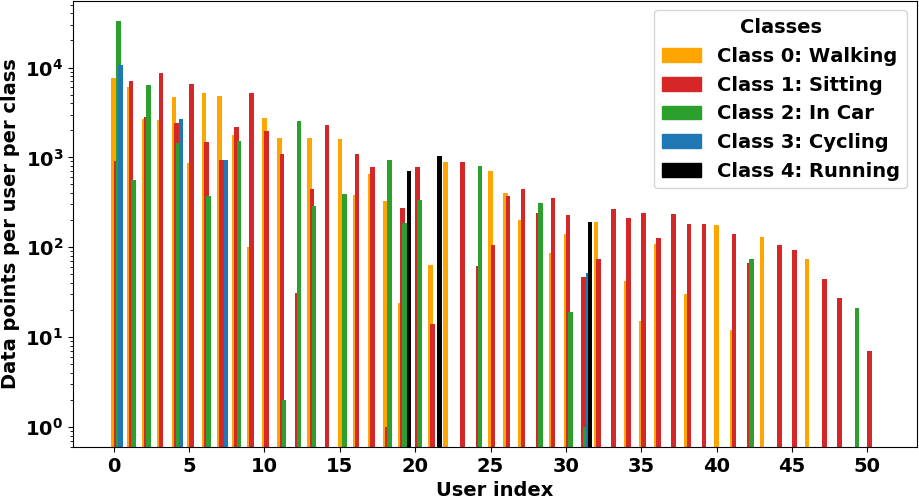} 
     \vspace{-2pt}
  \caption{Number of Data Points of Each Class for Each User} \vspace{-15pt}
  \label{fig_non_iid}
\end{figure}

\subsection{HAR-Wild Async Augmented Training} 

The performance of FL models is negatively affected by non-IID data distribution~\cite{konecny2016federated,zhao2018federated,DBLP:journals/corr/abs-1912-04977}, and we observed this to be true for HAR-Wild as well. Figure~\ref{fig_non_iid} shows the distribution of the dataset we collected for HAR-Wild. 
To address this problem, we leverage data augmentation training \cite{jeong2018communication} and tailor it to mitigate the distortion in computing gradients at client-side by balancing the client data with a small number of augmentation data samples without an undue computational cost.

The pseudo-code for HAR-Wild's asynchronous augmented learning is shown in Algorithm~\ref{alg_uniform_aug}. This algorithm is integrated in Algorithm~\ref{alg_async_fedavg} by replacing lines 7-12 from Algorithm~\ref{alg_async_fedavg} with the \Call{AugmentedGradients}{} procedure in Algorithm~\ref{alg_uniform_aug}. Before the whole training process starts, the FL Cloud Manager executes the procedure \Call{Init}{} (lines 1-3, Algorithm~\ref{alg_uniform_aug}), which first collects a small pool of random samples for each class that will be used for data augmentation (line 2). These data can be collected from a small number of volunteers or controlled users who share IID data with the FL Cloud Manager in FLSys. The augmentation data pool could also come from publicly available datasets. 
Then, the augmentation data pool $\mathcal{A}$ is delivered to each client (line 3). 
In each training round, each client (i.e., phone) randomly samples the augmentation data (line 8). Then, the sampled augmentation data $\mathcal{D_{A}}$ will be combined with the local data $\mathcal{D_{L}}$ (line 10, \Call{Concatenate}{$\mathcal{D_{A}}$, $\mathcal{D_{L}}$}) to compute the local gradients (lines 11-13, \Call{LocalTraining}{}). The local gradients are then sent to the cloud for the asynchronous average aggregation and model update (line 14).


In order to deliver the augmentation data to the clients (line 3), we consider two objectives: \textbf{(i)} privacy protection, and \textbf{(ii)} communication efficiency. One naive approach is to send data to augment the missing classes at the clients in each training round, since the local missing data can change over time. In this approach, the FL Cloud Manager needs to know which classes are missing for each client in each training round. This could increase the communication cost and significantly increase data privacy risk, since the cloud learns certain aspects of the user behavior based on the classes that miss data over time. To achieve both privacy protection and communication efficiency, the approach implemented in FLSys (Algorithm 2) first delivers the entire augmentation data to every client only once at the beginning of the training process. Then, the clients use only the data necessary to augment their missing data. The clients check the missing classes when they receive the data, and re-check every time they accumulate enough new data (the amount of new data is a model-specific configuration parameter).

\begin{algorithm}[t]
    \footnotesize
	\caption{HAR-Wild Asynchronous Augmented Learning~\label{alg_uniform_aug}} 
	\begin{algorithmic}[1]
	    \Procedure{Init}{$clients$}
    	    \State augmentation pool $\mathcal{A} \gets$ \Call{sampleAugmentData}{$clients$}
    	    \State \Call {deliverAugmentPool}{$\mathcal{A}$, $clients$}
	    \EndProcedure
	    \Procedure{AugmentedGradients}{Round $t$, Client $i$}
	        \State Augmentation data pool $\mathcal{A}$ 
	        \State Local data pool $\mathcal{L}_{i}$
            \State $\theta_{l} \gets \theta^{t}$
            \State augmentation data $\mathcal{D_{A}}$ = \Call{sampleAugmentData}{$\mathcal{A}$}
            \State local data $\mathcal{D_{L}}$ = \Call{sampleData}{$\mathcal{L}_{i}$}
            \State training data $\mathcal{D_{T}}$ = \Call{concatenate}{$\mathcal{D_{A}}$, $\mathcal{D_{L}}$}
            \For{batch $b \in \mathcal{D_{T}}$}
                \State $\theta_{l} \gets \theta_{l}-\eta \nabla \mathcal{L}(\theta_{l};b)$
            \EndFor
            \State $\Delta_{l} \gets \theta_{l} - \theta^t$
            \State \Call{uploadClientGradients}{$\Delta_{i}$}
	    \EndProcedure
	\end{algorithmic} 
\end{algorithm} %

\section{Evaluation} 
\label{sec:results}

The evaluation has two main goals: \textbf{(i)} Analyze the performance of the two FL models, HAR-Wild and sentiment analysis (SA) with different aggregators and DP mechanisms.  
\textbf{(ii)} Quantify the system performance of FLSys with HAR-Wild and SA on Android and AWS. In terms of system performance, we investigate energy efficiency and memory consumption on the phone, system tolerance to phones that do not upload local gradients, and FL aggregation scalability in the cloud. We also study the overall response time for third party apps that use FLSys on the phone. For model evaluation, we use Accuracy, Precision, Recall, and F1-score metrics. For system performance, we report execution time and memory consumption for both the phones and the cloud, and battery consumption on the phones. 

Most of the evaluation is done with HAR-Wild, which illustrates a typical FL model based on mobile sensing data. To demonstrate that FL works for different models, we also show results for the SA model. The rest of the section is organized as follows: Section~\ref{sec:eval-cent} compares HAR-Wild against baseline models and evaluates the effect of data augmentation, different aggregators, and advanced privacy mechanisms on HAR-Wild's performance. Section~\ref{sec:sa} describes the sentiment analysis (SA) model, used to demonstrate FLSys's support for different models, and shows its performance. Section~\ref{sec:eval-emul} shows the HAR-Wild performance over the FLSys prototype, in terms of model accuracy, fault tolerance, and scalability. Since we did not have enough phones for larger-scale experiments, we show these results using Android/Linux emulators to replay each user's data. Finally, Section~\ref{sec:eval-phone} presents results for HAR-Wild and SA over FLSys on two types of Android phones.

\begin{table}[] 
\centering 
\caption{Number of Samples in the Dataset for 51 Users} 
    \vspace{-5pt}
\resizebox{0.4\textwidth}{!}{
\begin{tabular}{|l|l|l|l|l|l|}
\hline
Type & \shortstack{Class 0 \\ Walking} & \shortstack{Class 1 \\ Sitting}  & \shortstack{Class 2 \\ In Car} & \shortstack{Class 3 \\ Cycling} & \shortstack{Class 4 \\ Running}  \\ \hline
Training  &  48855  &  51499  &  49185 & 14281 & 1920 \\ \hline
Testing  &  8514  &  8828  & 8595 & 2514 & 319   \\ \hline
\end{tabular}}
\label{table_dataset}    \vspace{-5pt}
\end{table}

\begin{table}[t]
\centering
\caption{Model Settings of HAR-W and Baselines} 
    \vspace{-5pt}
\label{table_model_settings}
\resizebox{0.5\textwidth}{!}{
\begin{tabular}{|c|c|c|}
\hline
Model & Optimizer & Other key parameters \\ \hline
\makecell{HAR-Wild \\(centralized)}  & Adam  & \makecell{LR=0.0005, dropout\_rate=0.4, batch\_size=1024 \\
Sampling: Same as class distribution} \\ \hline

\makecell{HAR-Wild\\ (sim-FL)}  & Adam  & \makecell{client\_LR=0.005, server\_LR=1.0, dropout\_rate=0.4, batch\_size=128,\\
Sampling: $[50,100]$ samples per class, $[15,30]$ augment samples per class} \\ \hline

\makecell{HAR-Wild\\ (sim-FL with \\additional\\aggregators)}  & Adam  & \makecell{client\_LR=0.005, server\_LR=1.0,  dropout\_rate=0.4, batch\_size=128 \\
 degree\_of\_adaptivity = 1, decay\_parameters = 0.1, 0.9\\
Sampling: $[50,100]$ samples per class, 
$[15,30]$ augment samples per class} \\ \hline

\makecell{HAR-Wild \\(sim-FL \\ with DP)}  & Adam  & \makecell{client\_LR=0.005, server\_LR=1.0,  dropout\_rate=0.4, batch\_size=256 \\
Sampling: $[50,100]$ samples per class, 
$[15,30]$ augment samples per class} \\ \hline

\makecell{HAR-Wild\\ (FLSys) } & Adam  & \makecell{client\_LR=0.005, server\_LR=1.0,  dropout\_rate=0.4, batch\_size=64 \\
Sampling: $[50,100]$ samples per class, 
$[15,30]$ augment samples per class} \\ \hline

\makecell{CNN-Ig \\(centralized) } & Adam  & \makecell{LR=0.0005, dropout\_rate=0.05, batch\_size=1024 \\
Sampling: Same as class distribution} \\ \hline

\makecell{BiLSTM\\ (centralized)}  & Adam  & \makecell{LR=0.0005, dropout\_rate=0.2, batch\_size=1024 \\
Sampling: Same as class distribution} \\ \hline

\end{tabular}} \vspace{-10pt}
\end{table}

\subsection{HAR-Wild Model Evaluation}  
\label{sec:eval-cent}
Table~\ref{table_dataset} shows the basic information of our collected dataset used for all HAR-Wild experiments. Some users have very limited numbers of labeled activities; thus, we select data from 51 users who labeled a reasonable amount of samples. 

\textbf{Comparison with Baseline Approaches.}
We perform centralized evaluation to assess HAR-Wild's utility compared to several baselines. Centralized training works as an upper bound performance for FL models. In addition, it allows us to fine-tune the model's hyper parameters. 
The evaluation includes three variants of HAR-Wild: \emph{HAR-W-32}, \emph{HAR-W-64}, and \emph{HAR-W-128}, which have the numbers of convolution-channels set to 32, 64, and 128. 
For comparison, we consider two baseline models: (1) \emph{Bidirectional LSTM} with 3-axial accelerometer data as input. This is a typical model for time-series data, and we fine-tune it based on grid-search of hyperparameters; and (2) The CNN-based models proposed by Ignatov~\cite{ignatov2018real}, with(\emph{CNN-Ig}) and without(\emph{CNN-Ig\_featureless}) additional features using the author's recommended settings in~\cite{ignatov2018real}. 
For a fair comparison, we used TensorFlow implementations for all models. 
Table~\ref{table_model_settings} shows all the hyper-parameters and model configurations. 

\begin{figure}[t]
  \centering
  \includegraphics[width=0.99\linewidth]{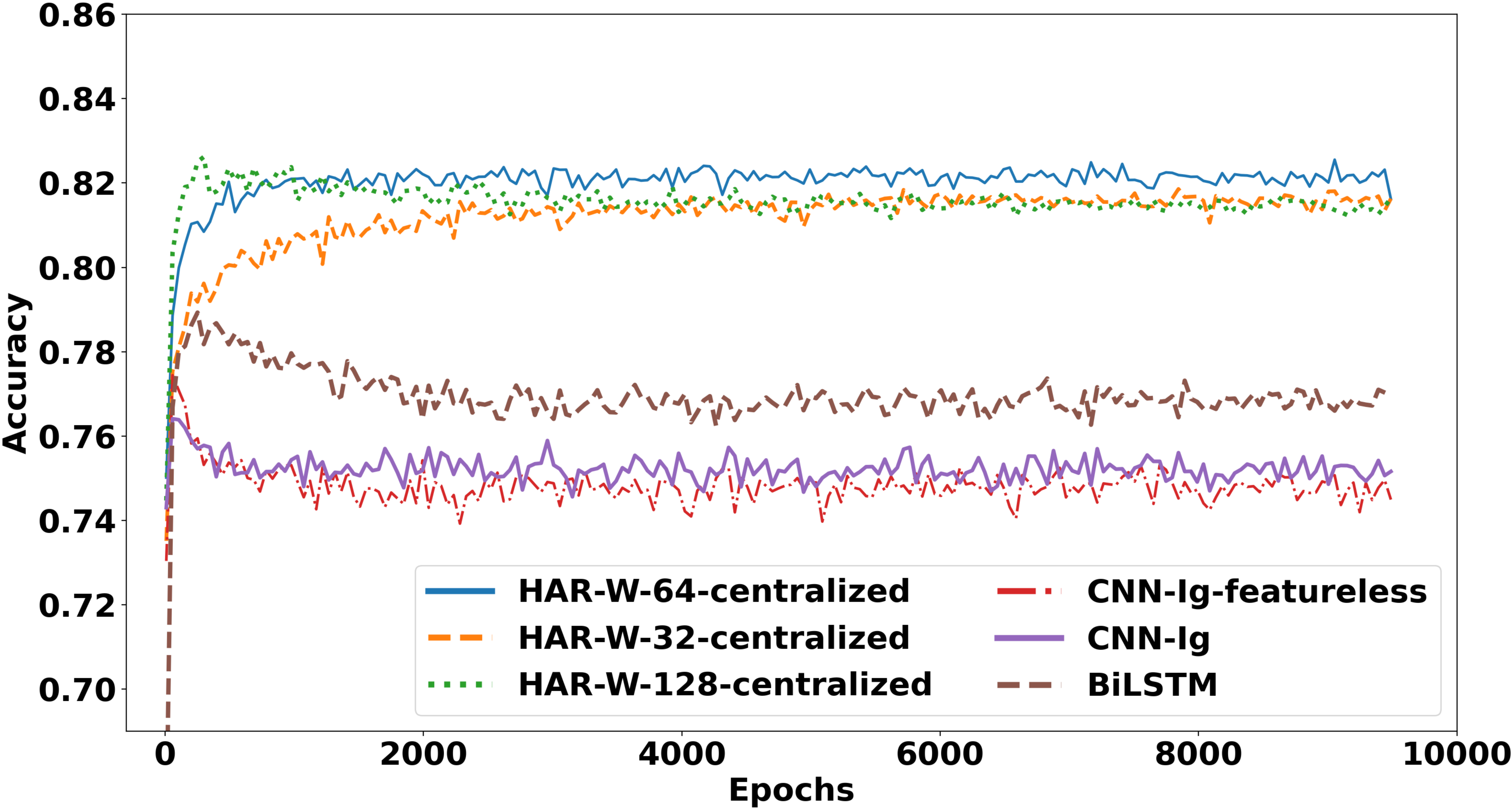} 
      \vspace{-5pt}
  \caption{Centralized training evaluation} 
  \label{fig_centralized}
  \vspace{-2.5pt}
\end{figure}

\begin{table}[t]
\centering
\caption{HAR-Wild Using Centralized and FL Training vs. Baselines: Macro-Model Performance}
\label{table_cent_model_compare}
\resizebox{0.45\textwidth}{!}{
\begin{tabular}{|l|l|l|l|l|}
\hline
Model & Accuracy & Precision & Recall  & F1-score \\ \hline
HAR-W-32-centralized  & 0.8186  & 0.8486 &  0.8360 & 0.8409  \\ \hline
HAR-W-64-centralized  & 0.8249  & 0.8512 & 0.8354 & 0.8428  \\ \hline
HAR-W-128-centralized &  0.8262  & 0.8529 &  0.8449 &  0.8484  \\ \hline
BiLSTM   &  0.7868 & 0.8074 &  0.7831 &  0.7941  \\ \hline
CNN-Ig   &  0.7639 & 0.7970 &  0.7715 &  0.7834  \\ \hline
CNN-Ig\_featureless  & 0.7708  & 0.8004 &  0.7779 &  0.7878  \\ \hline
HAR-W-64-fed-stock   & 0.5368  & 0.3828 & 0.3569 & 0.3190  \\ \hline
HAR-W-64-fed-uniform   & 0.7181  & 0.7464 & 0.7419 & 0.7378  \\ \hline
HAR-W-64-fed-yogi   & 0.7107  & 0.6865 &   0.7731 &   0.7130  \\ \hline
HAR-W-64-fed-adam   & 0.7072  &  0.6829 &   0.7592 &    0.7058  \\ \hline
HAR-W-64-fed-adagrad   &0.6691   &  0.6030  &  0.7429 &   0.6358  \\ \hline
\end{tabular}} \vspace{-10pt}
\end{table}

Figure~\ref{fig_centralized} shows that HAR-Wild models outperform the baseline approaches.
While the experiments run for up to 10,000 epochs to determine the performance upper bound, we observe the accuracy achieves acceptable performance after 1,000 epochs.
 On average, HAR-W-64 performs best and achieves 82.49\% accuracy compared with 78.68\%, 76.39\%, and 77.08\% of the BiLSTM, CNN-Ig and CNN-Ig-featureless. 
The results in Table~\ref{table_cent_model_compare} demonstrate that our HAR-Wild models also achieve the best performance in all the other metrics. 
Let us note that the absolute performance results may appear low when compared to HAR models run on data collected in controlled environment. This is because the data collected in the wild is noisier and non-IID.
Overall, HAR-W-64 (60,613 trainable weights) has the best trade-off among model accuracy, convergence speed, and model size, and we use it in all the following experiments for HAR-Wild. 

\textbf{Comparison of Different FL Versions of HAR-Wild.}
We also perform FL simulations to compare HAR-Wild's performance across three dimensions: (1) with and without data augmentation (2) with different aggregators (3) with and without advanced privacy mechanisms. Since the simulations are in TensorFlow, we can also compare the FL results with the centralized training results. In the simulated FL, we replay the data collected in the wild for each user.

In the following, the basic FL HAR-Wild model without data augmentation and without privacy mechanisms is called \emph{HAR-W-64-stock}. The model with data augmentation, but without privacy mechanisms, is called \emph{HAR-W-64-uniform}. The augmentation data, consisting of 640 samples of each class, is fixed and shared with all clients. 

The modular design of FLSys supports different FL aggregators. In addition to the standard FedAvg, we train the HAR-Wild model in FL with three aggregators designed to handle non-IID data~\cite{reddi2020adaptive}: FedYogi, FedAdam and FedAdagrad. To evaluate privacy protection in HAR-Wild, we apply the two types of  privacy-preserving mechanisms available in FLSys (described in Section~\ref{sec:design} and Appendix A): \emph{User-level DP} (\emph{User-DP}) and \emph{Local DP} (\emph{LDP}). We experiment with one User-DP mechanism proposed by~\cite{mcmahan-userDP} and five LDP mechanisms: \textit{BitRand}~\cite{BitRand}, \textit{Duchi}~\cite{duchi-LDP}, \textit{Piecewise}~\cite{piecewise-LDP}, \textit{Hybrid}~\cite{piecewise-LDP}, \textit{Three-Outputs}~\cite{three-outputs-LDP}. All the hyperparameters are provided in Table~\ref{table_model_settings}.

\begin{figure}[t]
  \centering
  \includegraphics[scale=0.205]{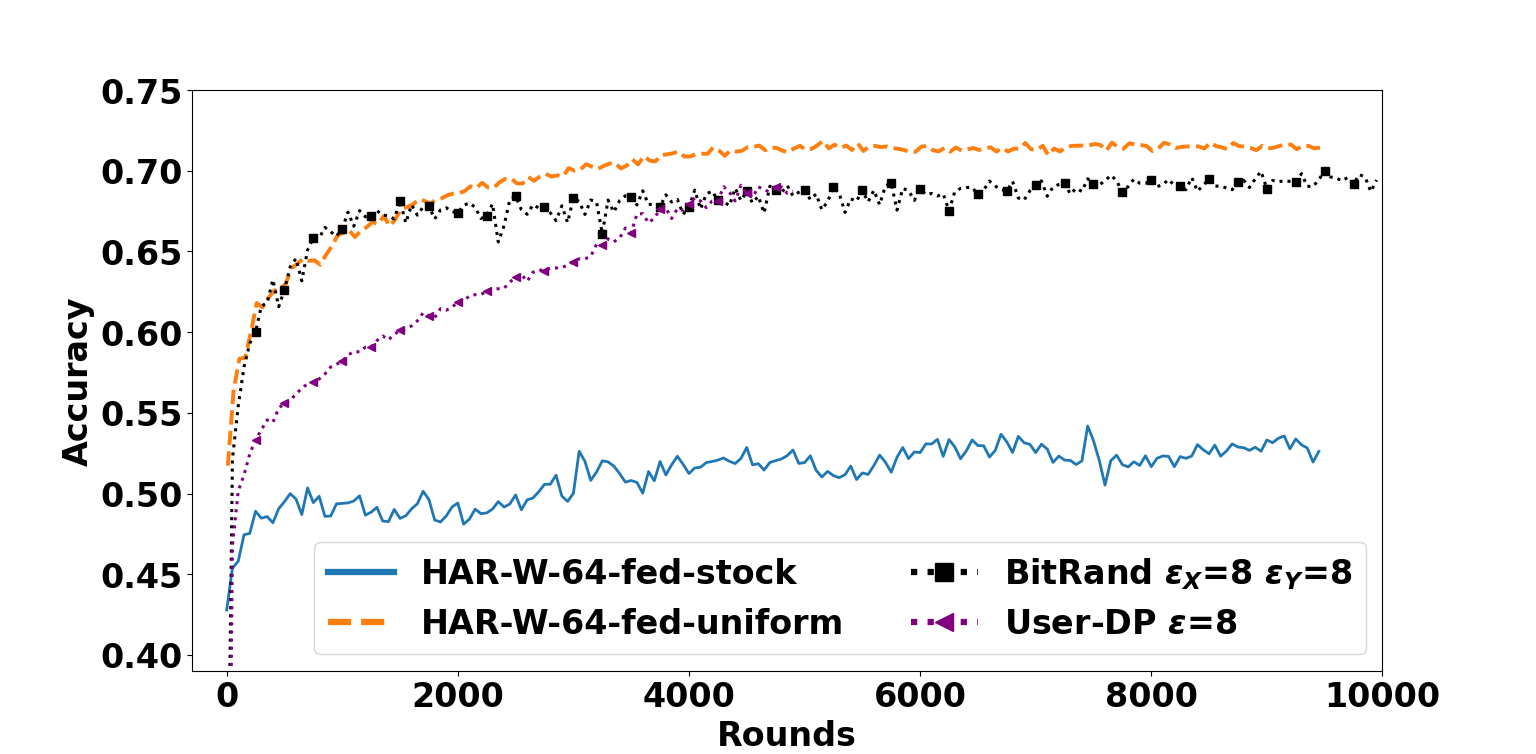}
  \caption{Comparison of FL HAR-Wild Versions, w/ and w/o Data Augmentation, and w/ and w/o Privacy Protection} 
  \label{fig_federated}
\end{figure}

\begin{table}[t]
\centering
\caption{Macro-model Performance for HAR-W-64-fed-uniform for Different Types of Privacy Protection Mechanisms and Different Parameters }
\label{table_DP_vs_nonDP}
\resizebox{0.46\textwidth}{!}{
\begin{tabular}{|l|l|l|l|l|l|}
\hline
DP Mechanism & Privacy Budget & Accuracy & Precision & Recall  & F1-score \\ \hline
Non-DP & $\varepsilon \to \infty$ & 0.7181  & 0.7464 &  0.7419 & 0.7378  \\ \hline
User-DP & $\varepsilon=2$ & 0.5399  & 0.5264 & 0.5797 & 0.5259  \\ \hline
User-DP & $\varepsilon=4$ & 0.5973  & 0.5603 & 0.6297 & 0.5502  \\ \hline
User-DP & $\varepsilon=8$ & 0.6970  & 0.6333 & 0.7264 & 0.6523  \\ \hline
BitRand & $\varepsilon_X = \varepsilon_Y=2$ & 0.4251 & 0.3667 & 0.3715 & 0.3277  \\ \hline
BitRand & $\varepsilon_X = \varepsilon_Y=4$ & 0.5193 & 0.4607 & 0.5110 & 0.4416 \\ \hline
BitRand & $\varepsilon_X = \varepsilon_Y=8$ & 0.6943 & 0.6885 & 0.7359 & 0.7031  \\ \hline
Duchi & $\varepsilon=2$ & 0.4846  & 0.4286 & 0.5233 & 0.4201  \\ \hline
Duchi & $\varepsilon=4$ & 0.5122  & 0.4307 & 0.4998 & 0.4360  \\ \hline
Piecewise & $\varepsilon=2$ & 0.4857  & 0.4086 & 0.4267 & 0.3944  \\ \hline
Piecewise & $\varepsilon=4$ & 0.5065  & 0.4245 & 0.4686 & 0.4222  \\ \hline
Hybrid & $\varepsilon=2$ & 0.4791  & 0.3961 & 0.3714 & 0.3714  \\ \hline
Hybrid & $\varepsilon=4$ & 0.5353  & 0.4521 & 0.4508 & 0.4431  \\ \hline
Three-Outputs & $\varepsilon=2$ & 0.2906  & 0.2662 & 0.2348 & 0.0192  \\ \hline
Three-Outputs & $\varepsilon=4$ & 0.2946  & 0.3288 & 0.2424 & 0.2386  \\ \hline
\end{tabular}} 
\end{table}


Table~\ref{table_cent_model_compare} shows the results for different FL versions of HAR-Wild.
HAR-W-64-fed-uniform (FedAvg with data augmentation) achieves 71.8\% accuracy, which is about 10\% less than the accuracy of the centralized-trained HAR-Wild. This is the cost of privacy-protection provided by FL. 

We tested FedYogi, FedAdam and FedAdagrad with and without data augmentation, and in both case they achieve comparable accuracy with FedAvg. Table~\ref{table_cent_model_compare} shows the results with data augmentation. Surprisingly, due to the noisy nature of HAR sensor data, the aggregators designed to handle non-IID data do not guarantee better performance than FedAvg. Therefore, the rest of the experiments will use FedAvg, which is the prevailing aggregator in FL.



\new{For privacy protection mechanisms, we train the HAR-W-64-fed-uniform model with the aforementioned DP mechanisms. Then, we evaluated the trade-offs between model utility and privacy budget for different versions of HAR-Wild with privacy mechanisms, as shown in Table 5.
As expected, the model utility decreases as privacy budget $\varepsilon$ tightens. From this table, we select the best User-DP model (i.e., the one with $\varepsilon=8$) and the best LDP model (i.e., BitRand with $\varepsilon_X=\varepsilon_Y=8$) in terms of accuracy, and compare them with the models with and without augmentation in Figure 5.
The results show that HAR-Wild with User-DP achieves a model accuracy of 69.70\%, which is just 2.11\% lower than the model without privacy protection. HAR-Wild with LDP (BitRand) achieves an accuracy of 69.43\%, which is just 2.38\% lower than the noiseless model. Note that our defense successfully prevents the server to reconstruct recognizable sensor signals and infer its associated ground-truth labels. One of the reasons is that it is more challenging to infer whether a time series of sensor signals belongs to a particular client than other domain applications. When using a tighter privacy budget, e.g., $\varepsilon_X = \varepsilon_Y = 4$ or $2$,  the gap between BitRand and Non-DP model becomes bigger. This is due to the fact that BitRand has not been designed for imbalanced data and cannot work well with significantly imbalanced data as our HAR dataset, especially when reducing the privacy budget $\varepsilon_Y$ for protecting the labels. Let us also emphasize that both privacy protection mechanisms offer rigorous privacy guarantees in FLSys without significant computational overhead.

The different aggregators and privacy preserving mechanisms also showcase how the modularity of FLSys can be used to easily exchange different implementations of a module.
}

\begin{table}[t]
\centering
\caption{SA Model Performance Per Class for Centralized and Federated Learning \label{table_SA_performance}} 
\resizebox{0.45\textwidth}{!}{
\begin{tabular}{|l|l|l|l|l|l|l|}
\hline
 & Class & Accuracy & Precision & Recall  & F1-score & Support \\ \hline
\multirow{2}{*}{CL} & negative  & \multirow{2}{*}{0.81} & 0.75   &   0.69   &   0.72   &   3159 \\\cline{2-2}\cline{4-7}
 & positive  &  & 0.84  &    0.88   &   0.86   &   5746
  \\\hline
\multirow{2}{*}{FL} & negative  & \multirow{2}{*}{0.79} & 0.73  &	0.64  	 &   0.68  &	3159
 \\\cline{2-2}\cline{4-7}
 & positive  &  & 0.81  &	0.87  &	    0.84  &	5746
  \\\hline

\end{tabular}} \vspace{-5pt}
\end{table}

\subsection{Sentiment Analysis (SA) Model Evaluation}  
\label{sec:sa}

FLSys is designed and implemented to be flexible, in the sense that training and inference of multiple models can run concurrently. On the server, different applications use independent AWS resources. On the phone, independent model trainers and inference runners are responsible for different applications. This subsection showcases the training performance of the SA model, a text analysis model that interprets and classifies the emotions (positive or negative) from text data. For example, with the inferred emotions of mobile users' private text data, a smart keyboard may automatically generate emoji to enrich the text before sending. 

We build the SA model for tweet data.
We use the FL benchmark dataset Sentiment140~\footnote{http://help.sentiment140.com/home}, which consists of 1,600,498 tweets from 660,120 users. We select the users with at least 70 tweets, and this sub-dataset contains 46,000+ samples from 436 users. Our SA model first extracts a feature vector of size 768 from each tweet with DistilBERT~\cite{sanh2019distilbert}. Then, it applies two fully connected layers with ReLU and Softmax activation, respectively, to classify the feature vector into positive or negative. The number of hidden states of the first fully connected layer is set to 128 to balance the convergence speed and model size. In the FL version of the model, 5\% of the users are used for data augmentation, and the rest of the users follow 4:1 train-test split. 

While the reference implementation associated with this benchmark dataset reached 70\% accuracy~\cite{caldas2019leaf} using 100 users with stacked LSTM in FL simulation, our SA model achieves superior performance, as shown in Table \ref{table_SA_performance}. Centralized learning achieves 81\% accuracy, while FL achieves 79\% accuracy (an acceptable drop). The FL version of this SA model will be further evaluated while running over FLSys on Android phones in Section~\ref{sec:eval-phone}.


\subsection{HAR-Wild over FLSys Emulation Performance}
\label{sec:eval-emul}
To evaluate the performance of HAR-Wild over the FLSys prototype at scale, we use Android emulation because we did not have enough phones for these experiments. Furthermore, since Android emulation is slow and costly, we run several larger-scale experiments with the same DL4J algorithms and functions in Linux, which is much faster. We train the model in these experiments for only 1,000 rounds because the simulation results showed that the accuracy is acceptable starting with this number of rounds.

All the phone components of the prototype, except for Data Collector and Data Preprocessor, run in the emulators. The cloud part of the prototype runs in AWS. The Android emulators run on top of virtual machines (VMs) in Google Cloud, as AWS does not support nested virtualization. We run 10 VMs in Google Cloud, and each VM has 16 vCPUs and 60GB memory. On each instance, we run 4 Android v10 emulators from AVD manager in Android Studio. Each emulator is loaded with 3 users' data files, and each file is sampled twice as different clients. In each round, each Android emulator participates in training on behalf of a few clients.
We set the deadline for the round in the FL Cloud Manager to 6 minutes.  

\begin{figure}[t!]
 \centering
 \includegraphics[scale=0.225]{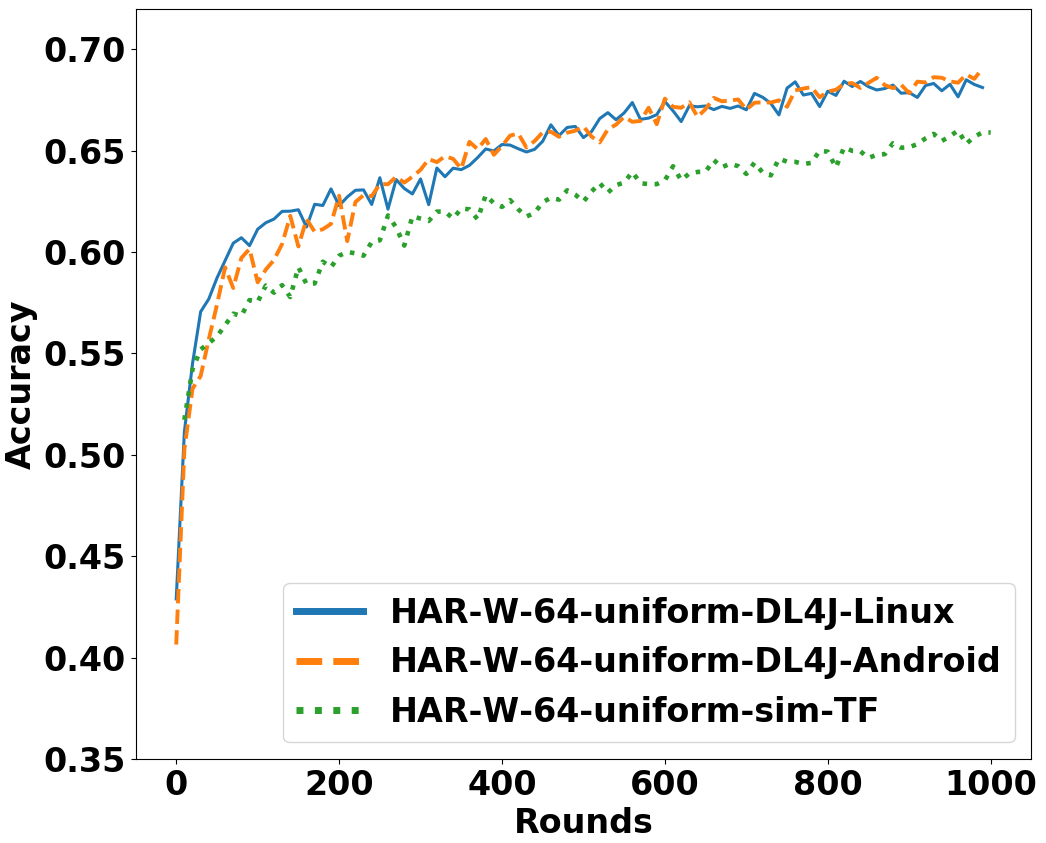}
\caption{HAR-Wild over FLSys Using Android/Linux Emulation} 
\label{fig_fed_sim_vs}
\end{figure}

\begin{table}[t!]
\centering
\caption{Performance Per Class of HAR-Wild over FLSys Using Android Emulation\label{table_fed_actual_performance}}
\resizebox{0.35\textwidth}{!}{
\begin{tabular}{|l|l|l|l|l|l|}
\hline
Class & Accuracy & Precision & Recall  & F1-score \\ \hline
 0  & \multirow{4}{*}{0.6907} & 0.7003 & 0.6628 & 0.6810 \\\cline{1-1}\cline{3-5}
 1  &  & 0.5922 & 0.8655 & 0.7032  \\\cline{1-1}\cline{3-5}
 2  &  & 0.8606 & 0.5443 & 0.6668 \\\cline{1-1}\cline{3-5}
 3  &  & 0.8324 & 0.6450 & 0.7268 \\\cline{1-1}\cline{3-5}
 4  &  & 0.6682 & 0.9028 & 0.7680 \\\hline
\end{tabular}} 
\end{table}

\begin{figure}[t!]
 \centering
 \includegraphics[scale=0.225]{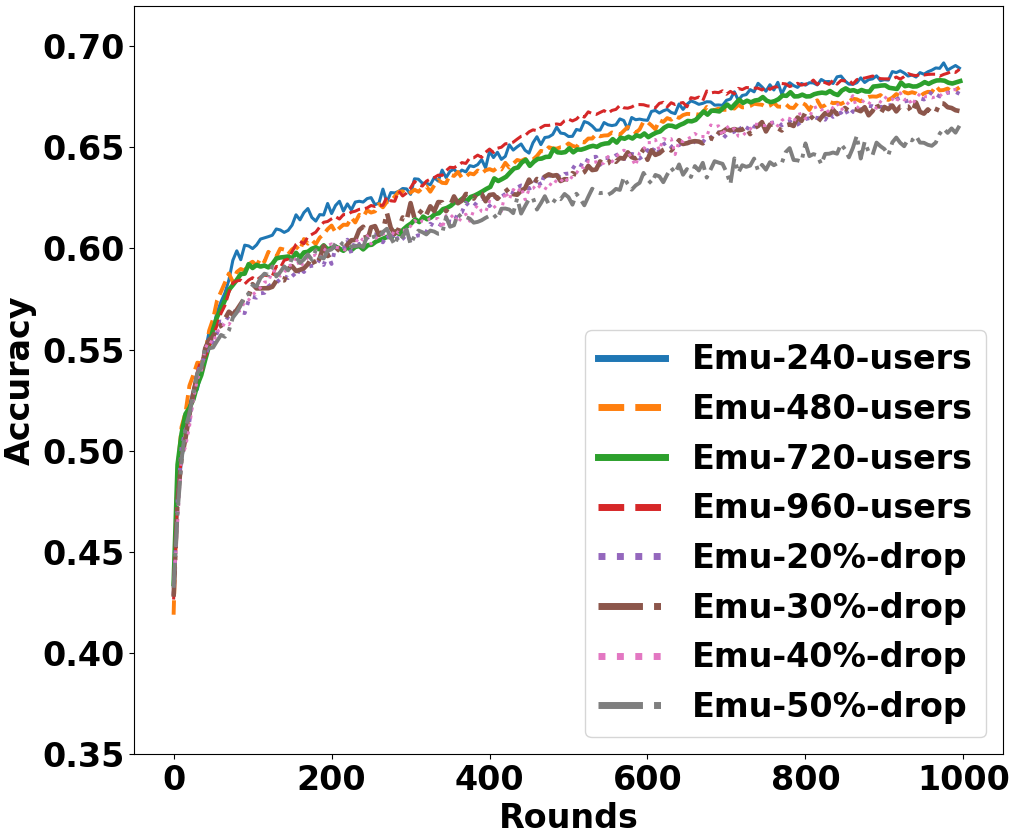}
 \caption{Linux Emulation of HAR-Wild over FLSys, while Varying Total Number of Users and Number of Users Dropping from Training} 
 \label{fig_acc_scale}
\end{figure}

\textbf{Accuracy.} Figure~\ref{fig_fed_sim_vs} shows that HAR-Wild with 64 clients emulation in both Android and Linux on FLSys achieve comparable accuracy with the simulated FL with TensorFlow, i.e., 69.07\%, 68.50\%, and 66.00\%. Table~\ref{table_fed_actual_performance} shows HAR-Wild's performance per class using FLSys and Android emulation. Although our data collected in the wild are inevitably unbalanced (Table~\ref{table_dataset}), every class performs reasonably well with F1-scores between 66.7\% and 76.8\%. Figure~\ref{fig_acc_scale} shows the results of HAR-Wild with higher number of clients (up to 960) using Linux emulations. The client data was over-sampled from the original 51 users.
HAR-Wild model achieves up to 69.17\% accuracy, and more clients help the model converge quicker with better performance.

\textbf{Fault Tolerance.}
In daily life, some clients may fail to upload a trained model to the FL Cloud Manager due to network or computation issues. This set of experiments verifies the fault tolerance of FLSys in terms of model performance as a given percentage of clients drop out randomly in each round. Figure~\ref{fig_acc_scale} shows the accuracy of HAR-Wild with up to 50\% clients dropping out randomly from 480 clients in each round. With 1,000 rounds of training, the accuracy is reduced by at most 3.11\%. This is a promising result showing that FLSys can tolerate reasonably large dropout rates during training. 

\begin{figure}[t!]
  \centering
  \includegraphics[width=0.75\linewidth]{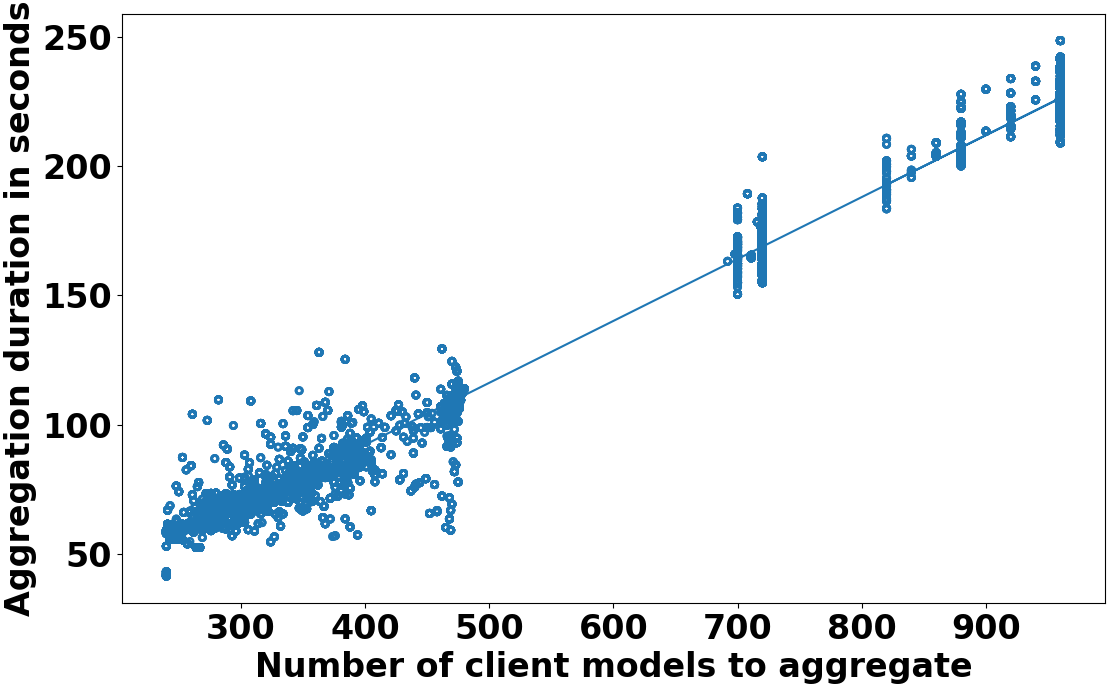}
  \caption{Aggregation time and participating clients} \vspace{-10pt}
  \label{fig_lambda_time_scale}
\end{figure}

\textbf{Scalability.}
As discussed in Section~\ref{sec:implementation}, computation and storage scale independently in the cloud for FLSys. This set of experiments verifies the scalability of FLSys across training rounds. The only FL function that may be computationally intensive in the cloud is the Model Aggregator. Figure~\ref{fig_lambda_time_scale} shows the Model Aggregator in AWS scales linearly with the number of participating clients. We also observe that the aggregation of 960 clients generally finishes in less than 4 minutes. By interpolating these results and given the current 15 minutes execution time limit of an AWS Lambda process~\cite{lambda}, the FLSys prototype (with single-threaded aggregator) can handle up to 3,600 clients, which is a sufficient number of clients, per training round. This number can be multiplied substantially by implementing both thread-level and process-level parallelization to handle real-world traffic volume.

Overall, the results for accuracy, fault-tolerance, and scalability demonstrate that FLSys and HAR-Wild can work well in real-life, where they are deployed on Android phones and the AWS cloud.
    \vspace{-5pt}

\subsection{FLSys Performance on Smart Phones}
\label{sec:eval-phone}
{\fontfamily{ptm}\selectfont

We benchmarked FLSys with HAR-Wild and SA on Android phones using a testing app to evaluate training and inference performance. We also assessed the resource consumption on the phones. We used three phones with different specs (Nexus 6P, Google Pixels 3 and 3a). The results demonstrate the on-device feasibility of FLSys, even for a low-end Nexus 6P phone, unveiled in 2015 and running Android 7. Since FLSys works well on such a low-end phone and people change their Android phones every 2-3 years on average~\footnote{https://www.statista.com/statistics/619788/average-smartphone-life/}, we expect FLSys to work well on most of today's phones.

\textbf{Training Performance.}
Table~\ref{training} shows the training time and the resource consumption on the phones. The training time is recorded by training 650 samples for 5 epochs for HAR-Wild, and 100 samples for 5 epochs for SA, which are the optimum scenarios determined in Section~\ref{sec:eval-emul}. Foreground training is done while leaving the screen on, and it uses the full single core capacity. It provides a lower bound for the training time. However, in reality, we expect training to be done in the background, either on battery or on charger.
As in practice, other apps or system processes working in background may interfere with training. We take 10 measurements for each benchmark, and report the mean and standard deviation.

\begin{table}[t!]
\centering
\caption{Training on Android Phones: Resource Consumption and Latency} 
    \vspace{-5pt}
\resizebox{0.5\textwidth}{!}{%
\begin{tabular}{|l|l|l|l|l|l|l|l|}
\hline
Model & Phone         
                & \shortstack{ Max \\RAM\\ Usage \\ (MB) } & \shortstack{ Foreground \\ Training \\ Time \\ Mean/SD\\ (min) } & \shortstack{ Background \\ Training \\ Time  on \\ Charger \\ Mean/SD \\(min) } & \shortstack{ Background \\Training \\ Time on \\ Battery \\ Mean/SD\\ (min) }
                 & \shortstack{Battery \\ Consumption \\ per Round \\ (mAh)}  & \shortstack{ Number \\of \\Training \\ Rounds \\ for Full\\ Battery }   \\ \hline
\multirow{3}{*}{HAR} 
& Nexus 6P     &219  & 4.95/0.94   & 39.10/26.10   & 45.34/24.31     & 35.10  & 98   \\ \cline{2-8}
& Google Pixel 3a     &156  & 1.23/0.01   & 3.94/0.04   & 85.82/33.07     & 9.72  & 308   \\ \cline{2-8}
& Google Pixel 3      &165  & 0.70/0.06  & 3.58/0.10   & 79.96/36.82  & 3.79  & 769   \\ \hline
\multirow{3}{*}{SA} 
& Nexus 6P       &139  & 1.62/0.08  & 5.04/0.13   & 29.79/17.13   & 7.94  & 435   \\ \cline{2-8}
& Google Pixel 3a     &128  & 0.33/0.005   & 0.84/0.006   & 25.42/5.72     & 2.02  & 1481   \\ \cline{2-8}
& Google Pixel 3      &136  & 0.22/0.002  & 0.76/0.02   & 24.19/8.12  & 0.76  & 3846   \\ \hline

\end{tabular}} 
\label{training}
\end{table} 

\begin{table}[t!]
\centering
\caption{Inference on Android Phones: Resource Consumption and Latency}
    \vspace{-5pt}
\resizebox{0.5\textwidth}{!}{%
\begin{tabular}{|l|l|l|l|l|l|l|l|}
\hline
Model & Phone          
                & \shortstack{ Max \\RAM\\ Usage \\ (MB) } & \shortstack{ Foreground \\ Inference \\ Time \\ Mean/SD\\ (millisecond) } & \shortstack{ Background \\ Inference \\ Time on \\ Charger \\ Mean/SD \\(millisecond) } & \shortstack{ Background \\Inference \\ Time on \\ Battery \\ Mean/SD \\(millisecond) }
                 & \shortstack{Battery \\ Consumption \\ per \\prediction \\ ($\mu$Ah)}  & \shortstack{  Millions \\ of \\inferences\\ for \\Full \\Battery }   \\ \hline
\multirow{3}{*}{HAR} 
& Nexus 6P       &161  & 54.65/16.36  & 1963.04/1540.29   & 7646.73/16349.49   & 4.49  & 0.77   \\ \cline{2-8} 
& Google Pixel 3a         &158  & 38.48/10.07   & 99.73/19.76   & 100.11/19.69     & 4.12  & 0.73 \\ \cline{2-8} 
& Google Pixel 3        &177  & 36.59/6.43  & 99.60/33.69   & 100.11/21.45   & 1.94  & 1.50   \\ \hline

\multirow{3}{*}{SA} 
& Nexus 6P       &114  & 19.66/6.06  & 20.10/20.04   & 20.25/28.11   & 3.35  & 1.03   \\ \cline{2-8}
& Google Pixel 3a         &108  & 11.90/3.71   & 20.65/4.45   & 19.58/3.93     & 2.3  & 1.30   \\ \cline{2-8}
& Google Pixel 3        &129  & 10.11/2.88  & 15.59/5.89   & 17.42/5.69   & 0.17  & 17.63   \\ \hline

\end{tabular}} 
\label{inference}
\end{table}

Training for one round is fast on the phones. The foreground training time on the more powerful phone, Pixel 3, is just 0.7 min for HAR-Wild, and 0.22 min for SA. The background training time on charger, which is the expected situation for FL training, is good for any practical situation. The phones experience a higher training time compared with the foreground case (completed one training round in less than 4 minutes). The background training time on battery is notably longer, since Android attempts to balance computation with battery saving. 

The results show training is also feasible in terms of resource consumption. The maximum RAM usage of the app is less than 165MB, and modern phones are equipped with sufficient RAM to handle it. While we did not perform experiments for battery consumption in the foreground (as this test was used just for a lower bound on computation time), we measured battery consumption for background training on battery. The phones could easily perform hundreds of rounds of training on a fully charged battery. It is worth noting that, typically, one round of training per day is enough, as the users need enough time to collect new data. 

\textbf{Inference Performance.}
The results in Table~\ref{inference} demonstrate that FLSys can be used efficiently by third-party apps. 
The inference time is measured within the third party testing app. Let us note that the inference is performed locally by the FL Phone Manager, without any network communication. Thus, the measured time consists of the inference computation time and the inter-process communication time. We continuously perform predictions for 30 minutes and report the average values. The inference time for the three scenarios on the third-party app, foreground, background on charger, and background on battery, follows a similar trend as training. FLSys and HAR-Wild/SA have reasonable resource consumption, which make them effective in practice. 

In addition to HAR and SA, many other applications may benefit from FLSys. For example, FL models are appropriate for privacy-sensitive image and video data collected on mobile devices. Existing research confirms that such models are feasible on resource-constrained mobile devices. For training, Mathur et al.~\cite{mathur2021device} demonstrated that training a 2-layer DNN classifier on top of a pre-trained MobileNet~\cite{howard2017mobilenets} on Android clients for the Office-31 dataset takes about 30 minutes to converge. For inference, we tested the inference time of MobileNet on 224*224 images, and it takes about 120ms for a single CPU thread. These numbers are comparable with our results on HAR and confirm that such models could run over FLSys.

\vspace{-5pt}


}
\section{Conclusions, Lessons Learned, and Future Work} 
\label{sec:conclusion}

This paper presented our experience with designing, building, and evaluating FLSys, an end-to-end federated learning system. FLSys was designed based on requirements derived from real-life applications that learn from mobile user data collected in the wild, such as human activity recognition (HAR). Compared with existing FL systems, FLSys balances model utility with resource consumption on the phones, tolerates client failures/disconnections and allows clients to join training at any time, supports multiple DL models that can be used concurrently by multiple apps, provides support for advanced privacy protection mechanisms, and  acts as a ``central hub'' on the phone to manage the training, updating, and access control of FL models used by different apps. We built a complete prototype of FLSys in Android and AWS, and used this prototype to demonstrate that FLSys is effective and efficient in practice in terms of model performance, privacy protection, resource usage, and latency. We believe FLSys can open the path toward creating an FL ecosystem of models and apps for privacy-preserving deep learning on mobile sensing data. In terms of actual deployment of FLSys in practice, we believe it can be offered as FL as a Service (FLaaS) by cloud providers.

Next, we report lessons learned and future work. The lessons learned are based on our experience with running FLSys on data collected in the wild from 50+ users over a 4-month period. Larger scales and longer periods are necessary for additional insights into system scalability and robustness, as well as model performance at scale.


\textbf{Build mechanisms to cope with non-IID data.} Since our data collection happened during the Covid-19 pandemic, we expected to see somewhat similar data from users who mostly stayed indoors. However, the data was non-IID, strengthening the idea that data collected in the wild will almost always be non-IID. 
A future work in FLSys is to provide support for model and data-specific augmentation and other approaches to cope with non-IID data.

\textbf{Beware the simulation pitfalls.} One common practice in FL simulations is to use the same instances/placeholders in memory for the different clients. Such simulations must carefully reset the instances for different clients to avoid any information leakage among clients, which can never happen in a real system. Our initial experiments showed unexpectedly different results between simulations and Android emulators with DL4J for the same settings. The first problem we discovered was that Batch Normalization (BN) is not supported in DL4J for specific data shapes. We implemented our own BN in DL4J, but the simulation results still did not match the experimental results. Finally, we realized that BN does not work well for FL (consistent with \cite{li2021fedbn}), but it does work in the simulations due to shared instances among the simulated clients. Thus, the FL models used in the reported experiments do not use BN. The second problem we noticed was that the Adam optimizer worked well for simulation, but not for the Android emulator experiments. This was also caused by shared instances accessed by all clients in the simulation. This should not happen in practice given privacy leakage through the shared instances. The lesson learned was that simulation may show better results than experiments with real systems for FL. Since most of FL papers in the literature are based on simulations, their results may suffer from similar problems with the ones described here. We believe FLSys offers an opportunity to test such FL models in real-life conditions.

\textbf{Balance mobile resources and model accuracy.} In the current FL literature, there are no results to show the FL models work well on mobile devices, while consuming a limited amount of resources on these devices (e.g., battery power, memory). A lesson that we understood early on is that FLSys will need to balance resource usage on mobiles with model accuracy. Therefore, FLSys used an asynchronous design in which policies on the mobile devices are evaluated to decide when it makes sense for the device to participate in training and consume resources. Our results show that good model accuracy can be achieved even when a significant number of mobile devices do not participate in training in order to save resources. Let us also note that real systems cannot expect to run the same number of rounds that we observe in simulations. For example, it is common to see 10,000 rounds in simulations. However, in real life, mobile devices may not train more than once a day due to both resource consumption and lack of enough new data. In such a situation, running 10,000 rounds will take over 27 years. Thus, models must be optimized for a realistic number of rounds. 

\textbf{Design for flexibility.} FLSys was designed for model flexibility on the phones from the beginning (i.e., allow apps to use multiple interchangeable models). However, we did not originally design for flexiblity in the cloud. At first, we used virtual machines in the cloud and durable cloud storage for all FL operations. However, when we analyzed scalability and performance issues, we realized that an FaaS solution and different types of storage are necessary. Therefore, we changed the design of the FLSys in the cloud to allow for different types of cloud platforms and storage options. Thus, FLSys can easily be ported to other cloud platforms beyond AWS.

\textbf{Future Work.} 
\new{FLSys provides a solid foundation to extend the privacy and security threat model and defense solutions. For example, the clients can be compromised to poison the federated training process by sending poisoning gradients to the server. To defend against such attacks, approaches such as robust aggregators, robust predictions, and certified guarantees for model classification can be integrated into the FLSys system. In fact, we can replace our current supported aggregators with robust aggregators in the FL Cloud Manager. In addition, robust predictions and certified guarantees can be integrated into the Model Runner in the FL Phone Manager. 

In addition to the work on security/privacy, we will add features to allow FLSys to support continuous data collection, which is what we expect to see in real-life scenarios. Finally, if FLSys is successful in creating an ecosystem of third-party apps and models, the long-term goal is to have it offered as an OS service, which improves efficiency and security.
}


\vspace{-10pt}
\bibliographystyle{plain}
\bibliography{sample-base, r1,r2,r3}

\begin{thebibliography}{10}

\bibitem{lambda}
{Amazon Web Services}.
\newblock {Lambda quotas}.
\newblock https://docs.aws.amazon.com/lambda/latest/dg/
  gettingstarted-limits.html, 2021.

\bibitem{anguita2013ucihar}
Davide Anguita, Alessandro Ghio, Luca Oneto, Xavier Parra, and Jorge~Luis
  Reyes-Ortiz.
\newblock A public domain dataset for human activity recognition using
  smartphones.
\newblock In {\em Esann}, volume~3, page~3, 2013.

\bibitem{beutel2021flower}
Daniel~J. Beutel, Taner Topal, Akhil Mathur, Xinchi Qiu, Javier
  Fernandez-Marques, Yan Gao, Lorenzo Sani, Kwing~Hei Li, Titouan Parcollet,
  Pedro Porto~Buarque de~Gusmão, and Nicholas~D. Lane.
\newblock Flower: A friendly federated learning research framework, 2021.

\bibitem{bonawitz2019towards}
K.~Bonawitz, H.~Eichner, W.~Grieskamp, D.~Huba, A.~Ingerman, V.~Ivanov,
  C.~Kiddon, J.~Kone{\v{c}}n{\`y}, S.~Mazzocchi, H.~B. McMahan, et~al.
\newblock Towards federated learning at scale: System design.
\newblock {\em arXiv preprint arXiv:1902.01046}, 2019.

\bibitem{10.1145/3133956.3133982}
Keith Bonawitz, Vladimir Ivanov, Ben Kreuter, Antonio Marcedone, H.~Brendan
  McMahan, Sarvar Patel, Daniel Ramage, Aaron Segal, and Karn Seth.
\newblock Practical secure aggregation for privacy-preserving machine learning.
\newblock In {\em Proceedings of the 2017 ACM SIGSAC Conference on Computer and
  Communications Security}, CCS '17, page 1175–1191, New York, NY, USA, 2017.
  Association for Computing Machinery.

\bibitem{caldas2019leaf}
Sebastian Caldas, Sai Meher~Karthik Duddu, Peter Wu, Tian Li, Jakub Konečný,
  H.~Brendan McMahan, Virginia Smith, and Ameet Talwalkar.
\newblock Leaf: A benchmark for federated settings, 2019.

\bibitem{carlini2020extracting}
N.~Carlini, F.~Tramer, E.~Wallace, M.~Jagielski, A.~Herbert-Voss, K.~Lee,
  A.~Roberts, T.~Brown, D.~Song, U.~Erlingsson, et~al.
\newblock Extracting training data from large language models.
\newblock {\em arXiv preprint arXiv:2012.07805}, 2020.

\bibitem{chai2020tifl}
Zheng Chai, Ahsan Ali, Syed Zawad, Stacey Truex, Ali Anwar, Nathalie Baracaldo,
  Yi~Zhou, Heiko Ludwig, Feng Yan, and Yue Cheng.
\newblock Tifl: A tier-based federated learning system.
\newblock In {\em Proceedings of the 29th International Symposium on
  High-Performance Parallel and Distributed Computing}, pages 125--136, 2020.

\bibitem{chavarriaga2013opportunity}
Ricardo Chavarriaga, Hesam Sagha, Alberto Calatroni, Sundara~Tejaswi Digumarti,
  Gerhard Tr{\"o}ster, Jos{\'e} del~R Mill{\'a}n, and Daniel Roggen.
\newblock The opportunity challenge: A benchmark database for on-body
  sensor-based activity recognition.
\newblock {\em Pattern Recognition Letters}, 34(15):2033--2042, 2013.

\bibitem{chen2016lstm}
Yuwen Chen, Kunhua Zhong, Ju~Zhang, Qilong Sun, and Xueliang Zhao.
\newblock Lstm networks for mobile human activity recognition.
\newblock In {\em 2016 International Conference on Artificial Intelligence:
  Technologies and Applications}. Atlantis Press, 2016.

\bibitem{9141436}
M.~Duan, D.~Liu, X.~Chen, R.~Liu, Y.~Tan, and L.~Liang.
\newblock Self-balancing federated learning with global imbalanced data in
  mobile systems.
\newblock {\em IEEE Transactions on Parallel \& Distributed Systems},
  32(01):59--71, jan 2021.

\bibitem{duchi-LDP}
John~C. Duchi, Michael~I. Jordan, and Martin~J. Wainwright.
\newblock Local privacy and statistical minimax rates.
\newblock In {\em 2013 IEEE 54th Annual Symposium on Foundations of Computer
  Science}, pages 429--438, 2013.

\bibitem{erlingsson2019amplification}
{\'U}.~Erlingsson, V.~Feldman, I.~Mironov, A.~Raghunathan, K.~Talwar, and
  A.~Thakurta.
\newblock Amplification by shuffling: From local to central differential
  privacy via anonymity.
\newblock In {\em Proceedings of the Thirtieth Annual ACM-SIAM Symposium on
  Discrete Algorithms}, pages 2468--2479, 2019.

\bibitem{FATE}
{FATE}.
\newblock {An Industrial Grade Federated Learning Framework}.
\newblock https://fate.fedai.org/, 2021.

\bibitem{9006000}
Z.~{Feng}, H.~{Xiong}, C.~{Song}, S.~{Yang}, B.~{Zhao}, L.~{Wang}, Z.~{Chen},
  S.~{Yang}, L.~{Liu}, and J.~{Huan}.
\newblock Securegbm: Secure multi-party gradient boosting.
\newblock In {\em 2019 IEEE International Conference on Big Data (Big Data)},
  pages 1312--1321, 2019.

\bibitem{gao2022feddc}
Liang Gao, Huazhu Fu, Li~Li, Yingwen Chen, Ming Xu, and Cheng-Zhong Xu.
\newblock Feddc: Federated learning with non-iid data via local drift
  decoupling and correction.
\newblock In {\em Proceedings of the IEEE/CVF Conference on Computer Vision and
  Pattern Recognition}, pages 10112--10121, 2022.

\bibitem{goodfellow2016deep}
Ian Goodfellow, Yoshua Bengio, Aaron Courville, and Yoshua Bengio.
\newblock {\em Deep learning}, volume~1.
\newblock MIT press Cambridge, 2016.

\bibitem{he2020fedml}
Chaoyang He, Songze Li, Jinhyun So, Xiao Zeng, Mi~Zhang, Hongyi Wang, Xiaoyang
  Wang, Praneeth Vepakomma, Abhishek Singh, Hang Qiu, Xinghua Zhu, Jianzong
  Wang, Li~Shen, Peilin Zhao, Yan Kang, Yang Liu, Ramesh Raskar, Qiang Yang,
  Murali Annavaram, and Salman Avestimehr.
\newblock Fedml: A research library and benchmark for federated machine
  learning, 2020.

\bibitem{he2016deep}
K.~He, X.~Zhang, S.~Ren, and J.~Sun.
\newblock Deep residual learning for image recognition.
\newblock In {\em IEEE conference on Computer Vision and Pattern Recognition},
  pages 770--778, 2016.

\bibitem{10.1145/3359789.3359824}
Zecheng He, Tianwei Zhang, and Ruby~B. Lee.
\newblock Model inversion attacks against collaborative inference.
\newblock In {\em Proceedings of the 35th Annual Computer Security Applications
  Conference}, ACSAC '19, page 148–162, New York, NY, USA, 2019. Association
  for Computing Machinery.

\bibitem{hernandez2019human}
Fabio Hern{\'a}ndez, Luis~F Su{\'a}rez, Javier Villamizar, and Miguel Altuve.
\newblock Human activity recognition on smartphones using a bidirectional lstm
  network.
\newblock In {\em 2019 XXII Symposium on Image, Signal Processing and
  Artificial Vision (STSIVA)}, pages 1--5. IEEE, 2019.

\bibitem{DBLP:journals/corr/HitajAP17}
Briland Hitaj, Giuseppe Ateniese, and Fernando P{\'{e}}rez{-}Cruz.
\newblock Deep models under the {GAN:} information leakage from collaborative
  deep learning.
\newblock {\em CoRR}, abs/1702.07464, 2017.

\bibitem{howard2017mobilenets}
Andrew~G Howard, Menglong Zhu, Bo~Chen, Dmitry Kalenichenko, Weijun Wang,
  Tobias Weyand, Marco Andreetto, and Hartwig Adam.
\newblock Mobilenets: Efficient convolutional neural networks for mobile vision
  applications.
\newblock {\em arXiv preprint arXiv:1704.04861}, 2017.

\bibitem{ignatov2018real}
Andrey Ignatov.
\newblock Real-time human activity recognition from accelerometer data using
  convolutional neural networks.
\newblock {\em Applied Soft Computing}, 62:915--922, 2018.

\bibitem{jeong2018communication}
Eunjeong Jeong, Seungeun Oh, Hyesung Kim, Jihong Park, Mehdi Bennis, and
  Seong-Lyun Kim.
\newblock Communication-efficient on-device machine learning: Federated
  distillation and augmentation under non-iid private data.
\newblock {\em arXiv preprint arXiv:1811.11479}, 2018.

\bibitem{DBLP:journals/corr/abs-1912-04977}
Peter Kairouz, H~Brendan McMahan, Brendan Avent, Aur{\'e}lien Bellet, Mehdi
  Bennis, Arjun~Nitin Bhagoji, Keith Bonawitz, Zachary Charles, Graham Cormode,
  Rachel Cummings, et~al.
\newblock Advances and open problems in federated learning.
\newblock {\em CoRR}, abs/1912.04977, 2019.

\bibitem{kim2021federated}
Muah Kim, Onur Günlü, and Rafael~F. Schaefer.
\newblock Federated learning with local differential privacy: Trade-offs
  between privacy, utility, and communication, 2021.

\bibitem{konecny2016federated}
Jakub Konečný, H.~Brendan McMahan, Daniel Ramage, and Peter Richtárik.
\newblock Federated optimization: Distributed machine learning for on-device
  intelligence, 2016.

\bibitem{kwapisz2011WISDM}
Jennifer~R Kwapisz, Gary~M Weiss, and Samuel~A Moore.
\newblock Activity recognition using cell phone accelerometers.
\newblock {\em ACM SigKDD Explorations Newsletter}, 12(2):74--82, 2011.

\bibitem{liu2020flame}
Ruixuan L., Yang C., Hong C., Ruoyang G., and Masatoshi Y.
\newblock {FLAME:} differentially private federated learning in the shuffle
  model.
\newblock {\em CoRR}, abs/2009.08063, 2020.

\bibitem{BitRand}
Phung Lai, Hai Phan, Li~Xiong, Khang~Phuc Tran, My~Thai, Tong Sun, Franck
  Dernoncourt, Jiuxiang Gu, Nikolaos Barmpalios, and Rajiv Jain.
\newblock Bit-aware randomized response for local differential privacy in
  federated learning, 2022.

\bibitem{li2020federated}
Tian Li, Anit~Kumar Sahu, Ameet Talwalkar, and Virginia Smith.
\newblock Federated learning: Challenges, methods, and future directions.
\newblock {\em IEEE Signal Processing Magazine}, 37(3):50--60, 2020.

\bibitem{li2021fedbn}
Xiaoxiao Li, Meirui Jiang, Xiaofei Zhang, Michael Kamp, and Qi~Dou.
\newblock Fedbn: Federated learning on non-iid features via local batch
  normalization.
\newblock {\em arXiv preprint arXiv:2102.07623}, 2021.

\bibitem{lin2013network}
Min Lin, Qiang Chen, and Shuicheng Yan.
\newblock Network in network.
\newblock {\em arXiv preprint arXiv:1312.4400}, 2013.

\bibitem{liu2019lifelong}
Boyi Liu, Lujia Wang, and Ming Liu.
\newblock Lifelong federated reinforcement learning: a learning architecture
  for navigation in cloud robotic systems.
\newblock {\em IEEE Robotics and Automation Letters}, 4(4):4555--4562, 2019.

\bibitem{liu2020fedsel}
R.~Liu, Y.~Cao, M.~Yoshikawa, and H.~Chen.
\newblock Fedsel: Federated sgd under local differential privacy with top-k
  dimension selection.
\newblock In {\em International Conference on Database Systems for Advanced
  Applications}, pages 485--501, 2020.

\bibitem{liu2020fedvision}
Yang Liu, Anbu Huang, Yun Luo, He~Huang, Youzhi Liu, Yuanyuan Chen, Lican Feng,
  Tianjian Chen, Han Yu, and Qiang Yang.
\newblock Fedvision: An online visual object detection platform powered by
  federated learning.
\newblock In {\em Proceedings of the AAAI Conference on Artificial
  Intelligence}, volume~34, pages 13172--13179, 2020.

\bibitem{luo2021no}
Mi~Luo, Fei Chen, Dapeng Hu, Yifan Zhang, Jian Liang, and Jiashi Feng.
\newblock No fear of heterogeneity: Classifier calibration for federated
  learning with non-iid data.
\newblock {\em Advances in Neural Information Processing Systems},
  34:5972--5984, 2021.

\bibitem{lyu2020towards}
L.~Lyu, Y.~Li, X.~He, and T.~Xiao.
\newblock Towards differentially private text representations.
\newblock In {\em Proceedings of the 43rd International ACM SIGIR Conference on
  Research and Development in Information Retrieval}, pages 1813--1816, 2020.

\bibitem{malekzadeh2021dopamine}
M.~Malekzadeh, B.~Hasircioglu, N.~Mital, K.~Katarya, M.~E. Ozfatura, and
  D.~G{\"u}nd{\"u}z.
\newblock Dopamine: Differentially private federated learning on medical data.
\newblock {\em arXiv preprint arXiv:2101.11693}, 2021.

\bibitem{marfoq2022personalized}
Othmane Marfoq, Giovanni Neglia, Richard Vidal, and Laetitia Kameni.
\newblock Personalized federated learning through local memorization.
\newblock In {\em International Conference on Machine Learning}, pages
  15070--15092. PMLR, 2022.

\bibitem{mathur2021device}
Akhil Mathur, Daniel~J Beutel, Pedro Porto~Buarque de~Gusmao, Javier
  Fernandez-Marques, Taner Topal, Xinchi Qiu, Titouan Parcollet, Yan Gao, and
  Nicholas~D Lane.
\newblock On-device federated learning with flower.
\newblock {\em arXiv preprint arXiv:2104.03042}, 2021.

\bibitem{mcmahan2017communication}
B.~McMahan, E.~Moore, D.~Ramage, S.~Hampson, and B.~A. y~Arcas.
\newblock Communication-efficient learning of deep networks from decentralized
  data.
\newblock In {\em Artificial Intelligence and Statistics}, pages 1273--1282,
  2017.

\bibitem{mcmahan-userDP}
H.~Brendan McMahan, Daniel Ramage, Kunal Talwar, and Li~Zhang.
\newblock Learning differentially private recurrent language models.
\newblock In {\em International Conference on Learning Representations}, 2018.

\bibitem{mugunthan2020privacyfl}
Vaikkunth Mugunthan, Anton Peraire-Bueno, and Lalana Kagal.
\newblock Privacyfl: A simulator for privacy-preserving and secure federated
  learning.
\newblock In {\em Proceedings of the 29th ACM International Conference on
  Information \& Knowledge Management}, pages 3085--3092, 2020.

\bibitem{murad2017deep}
Abdulmajid Murad and Jae-Young Pyun.
\newblock Deep recurrent neural networks for human activity recognition.
\newblock {\em Sensors}, 17(11):2556, 2017.

\bibitem{DBLP:conf/sp/NasrSH19}
Milad Nasr, Reza Shokri, and Amir Houmansadr.
\newblock Comprehensive privacy analysis of deep learning: Passive and active
  white-box inference attacks against centralized and federated learning.
\newblock In {\em 2019 {IEEE} Symposium on Security and Privacy, {SP} 2019, San
  Francisco, CA, USA, May 19-23, 2019}, pages 739--753. {IEEE}, 2019.

\bibitem{flare}
{Nvidia}.
\newblock {FLARE}.
\newblock https://nvidia.github.io/NVFlare/index.html, 2021.

\bibitem{pysyft}
{OpenMined}.
\newblock {PySyft}.
\newblock https://blog.openmined.org/tag/pysyft/, 2021.

\bibitem{8241854}
L.~T. {Phong}, Y.~{Aono}, T.~{Hayashi}, L.~{Wang}, and S.~{Moriai}.
\newblock Privacy-preserving deep learning via additively homomorphic
  encryption.
\newblock {\em IEEE Transactions on Information Forensics and Security},
  13(5):1333--1345, 2018.

\bibitem{reddi2020adaptive}
Sashank~J. Reddi, Zachary Charles, Manzil Zaheer, Zachary Garrett, Keith Rush,
  Jakub Kone{\v{c}}n{\'y}, Sanjiv Kumar, and Hugh~Brendan McMahan.
\newblock Adaptive federated optimization.
\newblock In {\em International Conference on Learning Representations}, 2021.

\bibitem{DBLP:journals/corr/abs-1812-06127}
Anit~Kumar Sahu, Tian Li, Maziar Sanjabi, Manzil Zaheer, Ameet Talwalkar, and
  Virginia Smith.
\newblock On the convergence of federated optimization in heterogeneous
  networks.
\newblock {\em CoRR}, abs/1812.06127, 2018.

\bibitem{sanh2019distilbert}
Victor Sanh, Lysandre Debut, Julien Chaumond, and Thomas Wolf.
\newblock Distilbert, a distilled version of bert: smaller, faster, cheaper and
  lighter.
\newblock {\em arXiv preprint arXiv:1910.01108}, 2019.

\bibitem{sarkar2020fedfocal}
Dipankar Sarkar, Ankur Narang, and Sumit Rai.
\newblock Fed-focal loss for imbalanced data classification in federated
  learning, 2020.

\bibitem{sun2020ldp}
L.~Sun, J.~Qian, and X.~Chen.
\newblock {LDP-FL}: Practical private aggregation in federated learning with
  local differential privacy.
\newblock {\em International Joint Conference on Artificial Intelligence},
  2021.

\bibitem{tfl}
{TensorFlow}.
\newblock {On-Device Training with TensorFlow Lite}.
\newblock
  \url{https://www.tensorflow.org/lite/examples/on_device_training/overview},
  2021.

\bibitem{tian2022flvoogd}
Yuhang Tian, Rui Wang, Yanqi Qiao, Emmanouil Panaousis, and Kaitai Liang.
\newblock Flvoogd: Robust and privacy preserving federated learning.
\newblock {\em arXiv preprint arXiv:2207.00428}, 2022.

\bibitem{8818446}
D.~{Verma}, G.~{White}, and G.~{de Mel}.
\newblock Federated ai for the enterprise: A web services based implementation.
\newblock In {\em 2019 IEEE International Conference on Web Services (ICWS)},
  pages 20--27, 2019.

\bibitem{10.1117/12.2519621}
Dinesh~C. Verma, Graham White, Simon Julier, Stepehen Pasteris, Supriyo
  Chakraborty, and Greg Cirincione.
\newblock {Approaches to address the data skew problem in federated learning}.
\newblock In Tien Pham, editor, {\em Artificial Intelligence and Machine
  Learning for Multi-Domain Operations Applications}, volume 11006, pages 542
  -- 557. International Society for Optics and Photonics, SPIE, 2019.

\bibitem{wagh2021dp}
S.~Wagh, X.~He, A.~Machanavajjhala, and P.~Mittal.
\newblock Dp-cryptography: marrying differential privacy and cryptography in
  emerging applications.
\newblock {\em Communications of the ACM}, 64(2):84--93, 2021.

\bibitem{piecewise-LDP}
Xiaokui Xiao, Yin Yang, Jun Zhao, Siu Hui, Hyejin Shin, Junbum Shin, and Ge~Yu.
\newblock Collecting and analyzing multidimensional data with local
  differential privacy.
\newblock 04 2019.

\bibitem{yang2019federated}
Q.~Yang, Y.~Liu, T.~Chen, and Y.~Tong.
\newblock Federated machine learning: Concept and applications.
\newblock {\em ACM Transactions on Intelligent Systems and Technology},
  10(2):1--19, 2019.

\bibitem{yang2018applied}
Timothy Yang, Galen Andrew, Hubert Eichner, Haicheng Sun, Wei Li, Nicholas
  Kong, Daniel Ramage, and Fran{\c{c}}oise Beaufays.
\newblock Applied federated learning: Improving google keyboard query
  suggestions.
\newblock {\em arXiv preprint arXiv:1812.02903}, 2018.

\bibitem{three-outputs-LDP}
Yang Zhao, Jun Zhao, Mengmeng Yang, Teng Wang, Ning Wang, Lingjuan Lyu, Dusit
  Niyato, and Kwok-Yan Lam.
\newblock Local differential privacy-based federated learning for internet of
  things.
\newblock {\em IEEE Internet of Things Journal}, 8(11):8836--8853, 2021.

\bibitem{zhao2018federated}
Yue Zhao, Meng Li, Liangzhen Lai, Naveen Suda, Damon Civin, and Vikas Chandra.
\newblock Federated learning with non-iid data.
\newblock {\em arXiv preprint arXiv:1806.00582}, 2018.

\end{thebibliography}

\vspace{-40pt}
\label{sec:bio}
%
\begin{IEEEbiography}[{\includegraphics[width=1in,height=1.25in,clip,keepaspectratio]{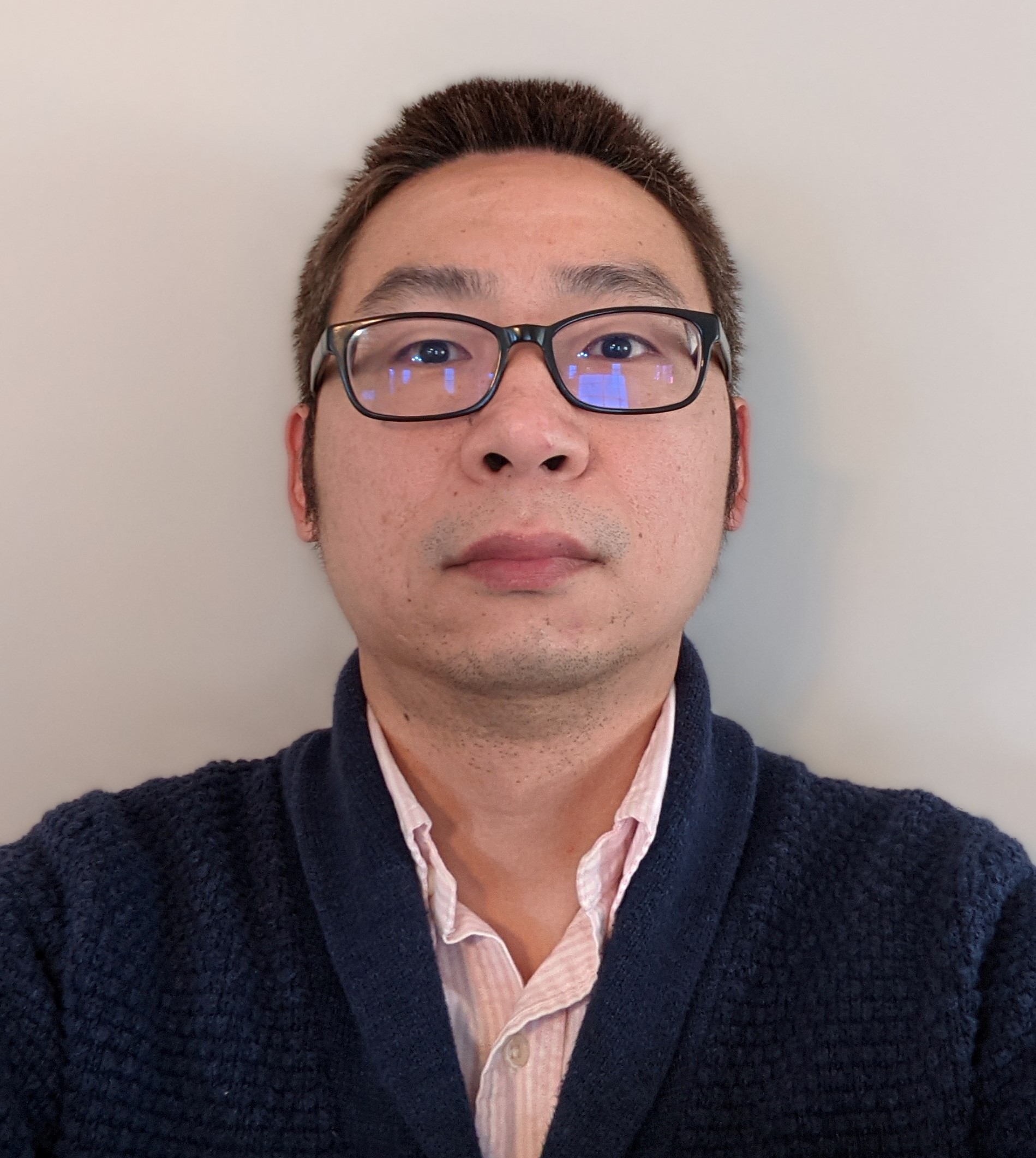}}]{Xiaopeng Jiang} is a PhD candidate in Computer Science at New Jersey Institute of Technology. His research interests
include deep learning systems and applications, mobile computing and sensing. Xiaopeng received an MS in Computer Science from NJIT in 2016.
\end{IEEEbiography}

\vskip -3\baselineskip
\begin{IEEEbiography}[{\includegraphics[width=1in,height=1.25in,clip,keepaspectratio]{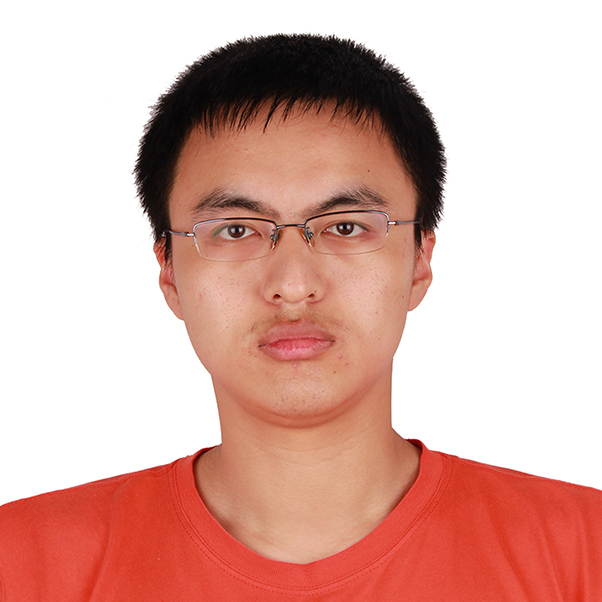}}]{Han Hu} received his Ph.D. degree in Information System from NJIT in December 2021. His research focuses on deep learning and federated learning for both industrial and social good applications. He has made publications in venues such as ICML, ICDM, IJCNN, SIGSPATIAL, ICHI, MedInfo, ICTAI and CSoNet.
\end{IEEEbiography}

\vskip -3\baselineskip
\begin{IEEEbiography}[{\includegraphics[width=1in,height=1in,clip,keepaspectratio]{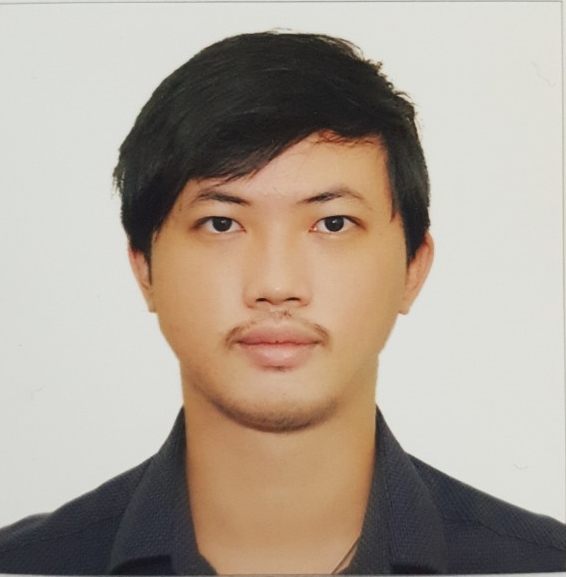}}]{Thinh On} is a first-year PhD candidate in Information Systems at New Jersey Institute of Technology. His research mainly focuses on deep learning and federated learning with an emphasis on privacy and security. He is also interested in investigating fairness, robustness, and trustworthiness of deep learning and federated learning.
\end{IEEEbiography}

\vskip -3\baselineskip
\begin{IEEEbiography}[{\includegraphics[width=1in,height=1.25in,clip,keepaspectratio]{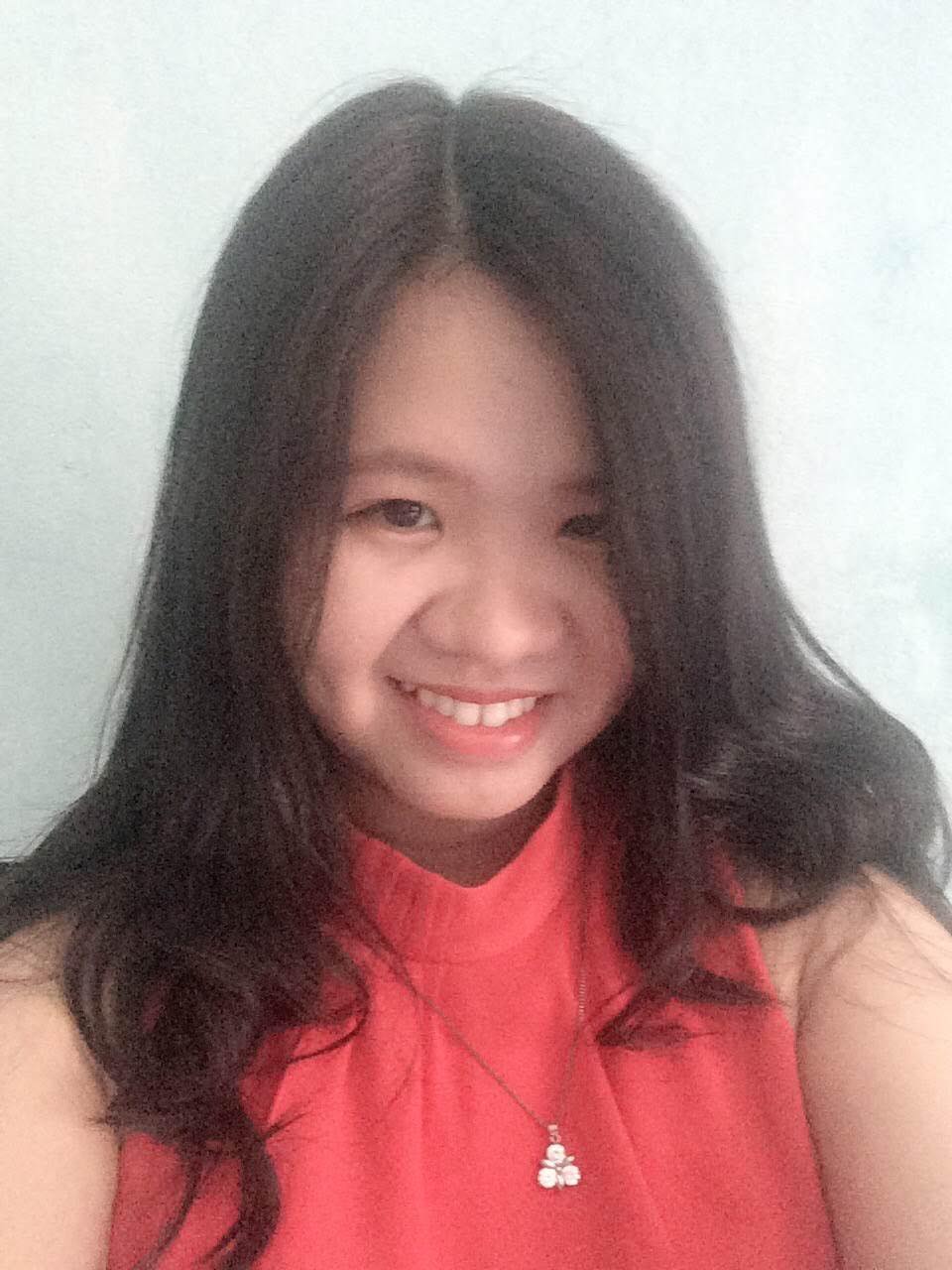}}]{Phung Lai} is a Ph.D. candidate in Information Systems at NJIT. Her research focuses on trustworthiness in machine learning and deep learning, including privacy preservation, explainability, robustness, and fairness techniques, with manifold applications such as natural language modeling, computer vision, social network analysis, finance, and healthcare. Phung is a holder of several patents in privacy preservation in NLP.
\end{IEEEbiography}

\vskip -3\baselineskip
\begin{IEEEbiography}[{\includegraphics[width=1in,height=1.25in,clip,keepaspectratio]{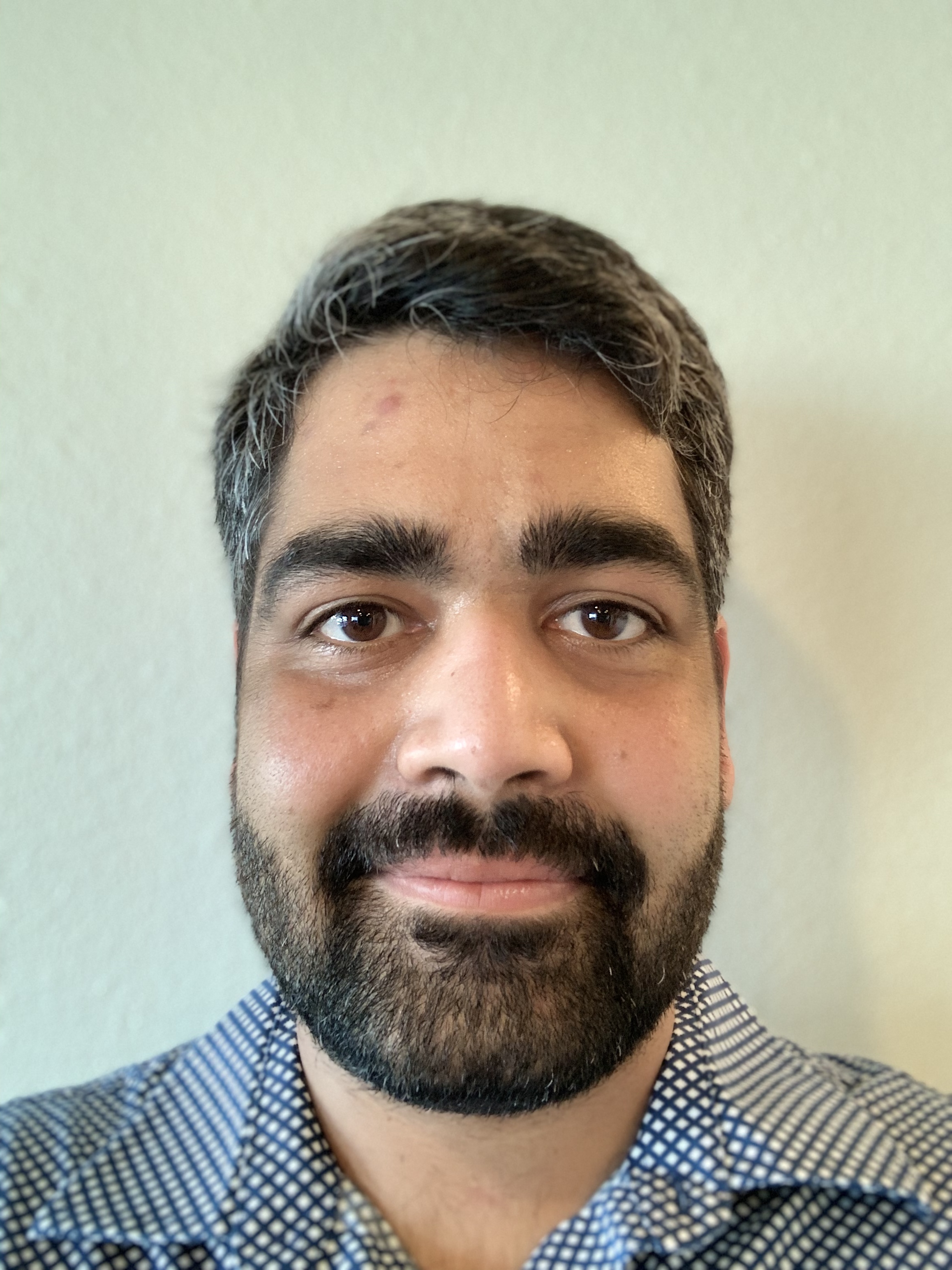}}]{Vijaya Datta Mayyuri} is a Principal Engineer at Qualcomm Incorporated. Vijaya has 15+ years of industry experience leading projects in various domains including 5G cellular connectivity, 4G/LTE, Wi-Fi, BLE , IOT, Medical Devices, and Machine Learning. Mr. Mayyuri received his master’s degree in Computer Science from The University of Texas at Dallas in 2006.
\end{IEEEbiography}

\vskip -3\baselineskip
\begin{IEEEbiography}[{\includegraphics[width=1in,height=1.25in,clip,keepaspectratio]{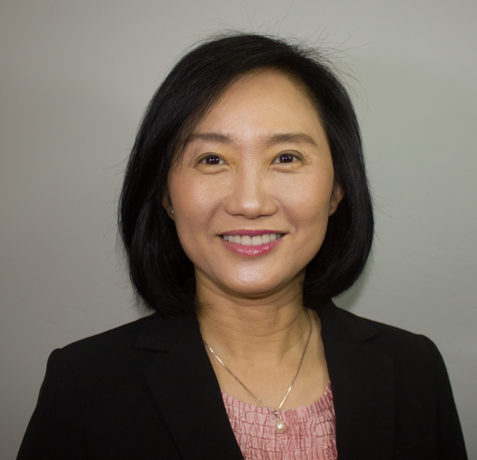}}]{An Chen} is a Vice President of Engineering at Qualcomm Incorporated with years of experience in the wireless industry.  She has done research and product development in advanced wireless, machine learning, mobile health, and IoT.  An is a prolific inventor with over 400 utility patents worldwide.   
An received her Ph.D. degree in electrical engineering from the University of California, San Diego.  She is a member of Phi Beta Kappa and Tau Beta Pi.
\end{IEEEbiography}

\vskip -3\baselineskip
\begin{IEEEbiography}[{\includegraphics[width=1in,height=1.25in,clip,keepaspectratio]{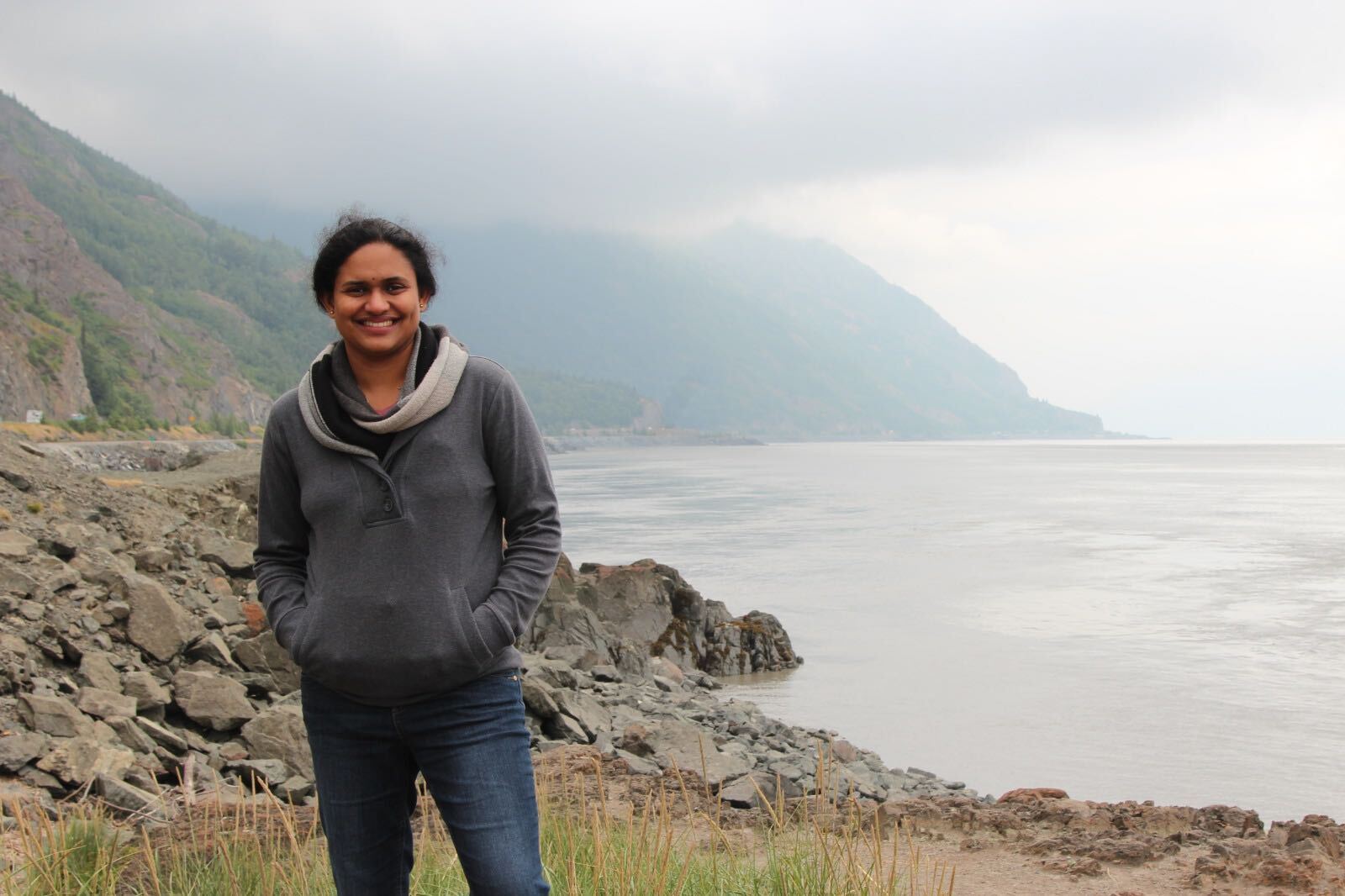}}]{Devu M. Shila} is the Founder and CEO of Unknot.id. She is a seasoned leader in building privacy-preserving mobile/wearable based human behavioral analytics products. She served as the Principal Investigator and product leader for advanced cyber security programs funded by DARPA, NSF, DoD, DHS S\&T, and DOE. Devu received her PhD in Computer Engineering from Illinois Institute of Technology in 2011.
\end{IEEEbiography}

\vskip -3\baselineskip
\begin{IEEEbiography}[{\includegraphics[width=1in,height=1.25in,clip,keepaspectratio]{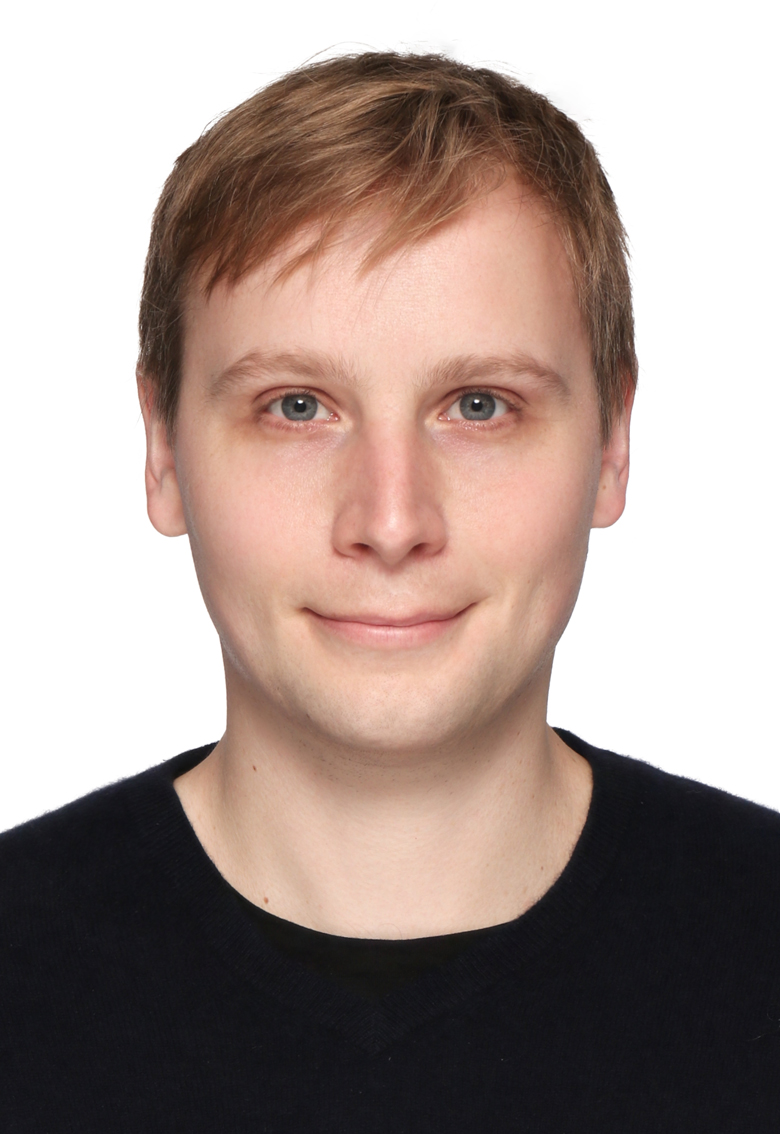}}]{Adriaan Larmuseau} earned his Ph.D. in Computer Science from Uppsala University in 2016. His research has focused on the cyber security of compilers, machine learning for patch management and genomic privacy. He has authored 12 published US \& WIPO patents.
\end{IEEEbiography}

\vskip -3\baselineskip
\begin{IEEEbiography}[{\includegraphics[width=1in,height=1.25in,clip,keepaspectratio]{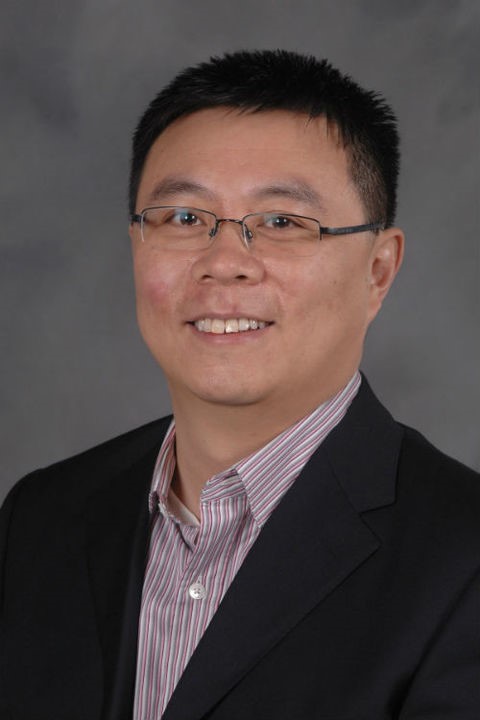}}]{Ruoming Jin} is a full professor in the department of computer science at Kent State University. His research areas include artificial intelligence/deep learning, recommendation systems, big data,  graph databases and health informatics. He has published over 150 research papers in these areas, most of them are in the top venues, such as KDD, ICDM, ICML, NIPS, SIGMOD, PVLDB etc. His research has been funded by NSF, NIH, SAMHSA, and industry partners.  He is the recipient of the NSF CAREER award.    

\end{IEEEbiography}

\vskip -3\baselineskip
\begin{IEEEbiography}[{\includegraphics[width=1in,height=1.25in,clip,keepaspectratio]{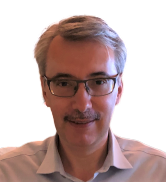}}]{Cristian Borcea} is a Professor in the Department of Computer Science at NJIT. He is also a Visiting Professor at the National Institute of
Informatics in Tokyo, Japan. His research interests include mobile computing and sensing, ad hoc and vehicular networks, distributed systems, and cloud computing. Borcea received his Ph.D. degree from Rutgers University. He is a member of the ACM and IEEE.
\end{IEEEbiography}

\vskip -3\baselineskip
\begin{IEEEbiography}[{\includegraphics[width=1in,height=1.25in,clip,keepaspectratio]{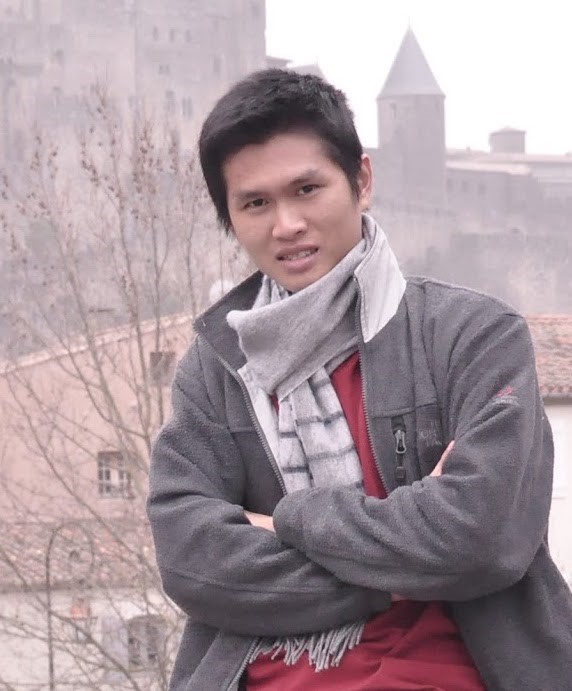}}]{NhatHai Phan} is an assistant professor in the Department of Data Science at NJIT. Hai's main topics of interests are privacy and security, machine learning, health informatics, social network analysis, and spatio-temporal data mining. Hai received his Ph.D. in Computer Science from the University of Montpellier 2 in 2013. Dr. Phan has published over 50 publications, with many of them were published at leading venues such as ICML, ECML, AAAI, IJCAI, IEEE ICDM, etc. His research is generously funded by NSF, Adobe Research, Qualcomm Incorporated, etc.
\end{IEEEbiography}

\end{document}